\documentclass{article}

\PassOptionsToPackage{numbers, compress}{natbib}
\usepackage[final]{neurips_2021}

\usepackage{textcomp}
\usepackage{xcolor}
\usepackage[utf8]{inputenc} % allow utf-8 input
\usepackage[T1]{fontenc}    % use 8-bit T1 fonts
\usepackage{hyperref}       % hyperlinks
\usepackage{url}            % simple URL typesetting
\usepackage{booktabs}       % professional-quality tables
\usepackage{amsfonts}       % blackboard math symbols
\usepackage{nicefrac}       % compact symbols for 1/2, etc.
\usepackage{microtype}      % microtypography
\usepackage{xcolor}
\usepackage{bbding}

\usepackage{graphicx}
\usepackage{subfigure}
\usepackage{wrapfig}
\usepackage{tabularx}
\usepackage{pifont}% http://ctan.org/pkg/pifont
\usepackage{amssymb}
\usepackage{amsmath}
\usepackage[noend]{algpseudocode}
\usepackage[ruled,vlined, linesnumbered]{algorithm2e}
\let\oldnl\nl% Store \nl in \oldnl
\newcommand{\nonl}{\renewcommand{\nl}{\let\nl\oldnl}}% Remove line number for one line
\newcommand{\Lagr}{\mathcal{L}}

\title{Learnable Fourier Features for Multi-Dimensional Spatial Positional Encoding}

\author{%
  Yang Li \\
  Google Research \\
  Mountain View, CA \\
  \texttt{liyang@google.com} \\
  \And
  Si Si \\
  Google Research \\
  Mountain View, CA \\
  \texttt{sisidaisy@google.com} \\
  \And
  Gang Li \\
  Google Research \\
  Mountain View, CA \\
  \texttt{leebird@google.com} \\
  \And
  Cho-Jui Hsieh \\
  UCLA \\
  Los Angeles, CA \\
  \texttt{chohsieh@cs.ucla.edu} \\
  \And
  Samy Bengio\thanks{Currently at Apple.} \\
  Google Research \\
  Mountain View, CA \\
  \texttt{bengio@gmail.com} \\
  
}

\begin{document}

\maketitle

\begin{abstract}
Attentional mechanisms are order-invariant. Positional encoding is a crucial component to allow attention-based deep model architectures such as Transformer to address sequences or images where the position of information matters. In this paper, we propose a novel positional encoding method based on learnable Fourier features. Instead of hard-coding each position as a token or a vector, we represent each position, which can be multi-dimensional, as a trainable encoding based on learnable Fourier feature mapping, modulated  with  a  multi-layer perceptron. The representation is particularly advantageous for a spatial multi-dimensional position, e.g., pixel positions on an image, where $L_2$ distances or more complex positional relationships need to be captured. Our experiments based on several public benchmark tasks show that our learnable Fourier feature representation for multi-dimensional positional encoding outperforms existing methods by both improving the accuracy and allowing faster convergence.

%{\color{blue}(Cho: some potential abation studies: a) Compare sin/consine position encoding + MLP  v.s. MLP only v.s. learnable Fourier features + MLP v.s. random  Fourier features + MLP; this can demonstrate whether learnable Fourier features are important) b) The results with/without MLP, show MLP is also important. }

\end{abstract}

\section{Introduction}
\label{section:introduction}
%Attention is important
%Positional encoding
%The issue with positional encoding
%our approach
%our contribution

Attentional mechanisms are a central component in many deep architectures~\citep{bahdanau2014neural, luong-etal-2015-effective}, which allow a model to selectively focus on specific information in the context. Transformer \citep{DBLP:journals/corr/VaswaniSPUJGKP17} and its many variants, such as~\citep{pmlr-v80-parmar18a, DBLP:journals/corr/VaswaniSPUJGKP17,kitaev2020reformer,carion2020endtoend}, which are solely based on attentional mechanisms, have advanced the state of the art on many tasks that involve data with inherent temporal and spatial orders, e.g., machine translation \citep{DBLP:journals/corr/VaswaniSPUJGKP17}, image generation \citep{kitaev2020reformer}, and object detection~\citep{carion2020endtoend}.

In contrast to recurrent~\citep{lstm, DBLP:journals/corr/abs-1808-03314,NEURIPS2019_383beaea} or convolutional architectures~\citep{cnn}, which automatically capture the ordinal information as computation progresses based on sequential or spatial dependencies, attentional mechanisms are order invariant. It allows a model to directly access information at an arbitrary position in a sequence or space. The lack of ordinal information in the model is not an issue when attentional mechanisms are combined with a recurrent or convolutional architecture~\citep{bahdanau2014neural, luong-etal-2015-effective}. However, it is crucial for Transformer-alike models where the entire model is built based on attentional mechanisms.

To capture positional information in the data, e.g., the token position in a sentence or the pixel coordinates in an image, \textit{positional encoding} has been introduced~\citep{pmlr-v70-gehring17a,DBLP:journals/corr/VaswaniSPUJGKP17}, where a position in a one or two-dimensional space is mapped to a vector space by either learning or heuristics-based approaches. The representation of an input, by combining both its positional encoding and content representation, e.g., word embeddings, then participates in downstream computation for attentional mechanisms.
The original Transformer model uses a fixed sinusoidal encoding with predefined wavelengths~\cite{DBLP:journals/corr/VaswaniSPUJGKP17}. However, the predefined features lack flexibility and may not capture important position information in a task-dependent manner.  To encode positions in a more flexible and data-driven way,  position embedding approaches (e.g., one used in BERT~\cite{devlin-etal-2019-bert}) introduce trainable embedding vectors for each (absolute or relative) position. Unfortunately, 
this data-driven approach comes at the cost of introducing a large amount of extra learnable parameters proportional to sequence lengths times the hidden dimension size. Moreover, it is non-trivial to apply position embedding to problems with variable sequence lengths. 

%Both of these position encoding approaches have been generalized to multi-dimensional positions. For instance, \cite{pmlr-v80-parmar18a,carion2020endtoend,visiontransformer} use multi

In this paper, we consider the problem of designing a position encoding for multi-dimensional spatial positions, such as pixel positions in an image or object bounding boxes in a spatial structure such as UIs. Existing methods typically use sinusoidal position encoding with hand-crafted frequencies or learned embedding to encode each dimension independently and then combine the resulting vector representations via concatenation, e.g., ~\citep{pmlr-v80-parmar18a,carion2020endtoend,visiontransformer}. Unfortunately, these approaches, by concatenating the representation of each dimension, are not effective to capture desired positional similarity on an image, such as $L_2$ distance or more complex positional relationships. While embedding-based approaches have the potential to learn complex positional relationships, since the number of unique positions grows exponentially to the input dimension, the approach incurs large overhead in 2D and could be infeasible scaling to a higher dimensional space. In addition, special treatments are needed to adjust the learned position embedding when the test image sizes differ from training, such as bicubic interpolation used in DeiT~\cite{touvron2020training} or Vision Transformer~\cite{visiontransformer}. To avoid these special adjustments, it is an important for positional encoding to handle unseen positions.

%Existing positional encoding methods are effective for addressing one-dimensional sequence problems, e.g., machine translation. For multi-dimensional positions, e.g., pixel positions in an image involving both vertical and horizontal dimensions, existing methods typically encode each dimension independently and then combine the resulting vector representations via concatenation, e.g., ~\citep{pmlr-v80-parmar18a,carion2020endtoend,visiontransformer}. As we will show in the following section, this approach does not capture desired positional similarity on an image. In general, this approach of dimensional-independent encoding followed concatenation has drawbacks for representing 2D or even higher-dimensional positions where L2 distances or more complex positional relationships are desired. 

The main contributions of our work are as follows. We design a novel positional encoding method that learns a function to map multi-dimensional positions into a vector space. The function extracts position information based on a set of Fourier features and passing them to an MLP. 
%Fourier features~\citep{RFF07,Tewari18}, which can represent 2D or higher-dimensional positions in a vector space. 
The encoding function is \textit{learnable} %representation is shift-invariant 
and is initialized in such a way that the inner products of our positional encodings approximate Euclidean distances. 
The inductive bias can be desirable in a 2D or higher-dimensional space and by learning from the data, the representation can be adapted to a specific problem. Since our method learns an encoding function instead of embedding vectors for each position, it is naturally \textit{inductive} and can handle test samples with arbitrary length. 
Our method is \textit{parameter-efficient}, in the sense that the number of parameters do not grow with sequence length. 
To allow complex positional relationships, our representation is also \textit{composable} by encoding each subset of dimensions, in a multi-dimensional space, using a shared learnable Fourier features. %Some important properties of our method are summarized in Table \ref{tab:pe_methods_compare}. 
We evaluate our method on a number of tasks where Transformer-based models have been used for problems with multi-dimensional positions, including image generation~\citep{kitaev2020reformer}, object detection~\citep{carion2020endtoend} and image classification~\citep{visiontransformer}, which all involve 2D positions (vertical and horizontal) in images. We also evaluate our method on natural language generation in graphical user interfaces, which involve modeling a sparse spatial structure of UI objects on the screen, where each object is characterized by 4-coordinate values (top, left, bottom, and right)~\citep{li-etal-2020-widget}. These experiments show that our positional encoding method consistently outperforms existing methods by both improving accuracy and accelerating learning.

%The main contributions of our work are as follows. We design a novel positional encoding method based on Fourier features~\citep{RFF07,Tewari18}, which can represent 2D or higher-dimensional positions in a vector space. Our \textit{learnable} representation is shift-invariant and is initialized in such a way that the inner products of our positional encodings approximate Euclidean distances. The \textit{inductive bias} can be desirable in a 2D or higher-dimensional space and by learning from the data, the representation can be adapted to a specific problem. Our method is \textit{parameter-efficient}, in the sense of the number of parameters do not grow with sequence length. 
%To allow complex positional relationships, our representation is also \textit{composable} by encoding each subset of dimensions, in a multi-dimensional space, using learnable Fourier features. We evaluate our method on a number of tasks where Transformer-based models have been used for problems with multi-dimensional positions, including image generation~\citep{kitaev2020reformer}, object detection~\citep{carion2020endtoend}, and image classification~\citep{visiontransformer} which all involve 2D positions in images. We also evaluate our method on natural language generation in graphical user interfaces, which involve modeling UI objects on the screen with 4-dimensional positions~\citep{li-etal-2020-widget}. These experiments show that our positional encoding method consistently outperform existing methods by both improving accuracy and accelerating learning.

\section{Background}
\label{section:background}
%We review the positional encoding in the context of attention-based models (Section~\ref{section:attention-models}) and describe the issues with existing approaches (Section~\ref{section:positional} and~\ref{section:multidim}). We then review random Fourier features (Section~\ref{section:fourier}), which is a foundation for our method for positional encoding that is to be introduced in Section~\ref{section:model}.

%\begin{table*}[h]
%\small
%\centering
%\begin{tabularx}{0.9 \textwidth}{l|c|c|c|c}
%    \hline
%    {Positional Embedding}
%    & Inductive &Learnable &Parameter Efficient& Multi-dimensional \\
%    \hline  \hline
%    Sinusoidal\cite{DBLP:journals/corr/VaswaniSPUJGKP17} & \checkmark&\xmark &\checkmark & \checkmark \\
%   \hline 
%   Learned Embedding\cite{devlin-etal-2019-bert} & \xmark&\checkmark &\xmark & \checkmark\\
%   \hline
%      FLOATER\cite{Liu2020LearningTE}& \checkmark&\checkmark &\checkmark & \xmark\\
%   \hline
%      Our method& \checkmark&\checkmark &\checkmark & \checkmark\\
%   \hline
%\end{tabularx}
%\vspace{-5pt}
%  \caption{Comparison among different position encoding methods. 1. Inductive: the ability to handle sequences longer than any sequence seen in the training time.
%2. Data-Driven: the position encoding should be learnable from the data.
%3. Parameter Efficient: number of trainable parameters introduced by the encoding should be limited
%to avoid increased model size, which could hurt generalization. 4. Multi-dimensional: whether can be used for multi-dimensional settings. }
%  \label{tab:pe_methods_compare}
%\end{table*}
%\vspace{-5pt}

\subsection{Positional Encoding}
\label{section:positional}
In Transformer models, the self-attentional mechanism determines the strength between each pair of items based on the dot product similarity of their vector representations, which are derived from an item's content embedding and positional encoding~\cite{DBLP:journals/corr/VaswaniSPUJGKP17} (Appendix~\ref{section:attention-models}). Although positional encoding (PE) does not function alone in determining the attention strength, the benefit of having the inductive bias of positional relevance in the PE is evidenced by the success of the sinusoidal positional encoding originally proposed in Transformer~\cite{DBLP:journals/corr/VaswaniSPUJGKP17} (Equation~\ref{eq:sin-cos}).
\begin{equation}
    \label{eq:sin-cos}
        PE(p,2d) = \sin{\frac{p}{10000^{2d/D}}}; 
        PE(p,2d+1) = \cos{\frac{p}{10000^{2d/D}}}
\end{equation} 

which encodes a scalar position, $p$, using sinusoidal functions with different constant frequencies for each dimension, $d$, of a $D$-dimensional encoding vector. The dot product of this encoding representation naturally captures positional similarity in a 1D sequence in a parameter-free fashion. 

\begin{wrapfigure}{r}{0.45\textwidth}
\centering
\centerline{\includegraphics[width=0.4\columnwidth]{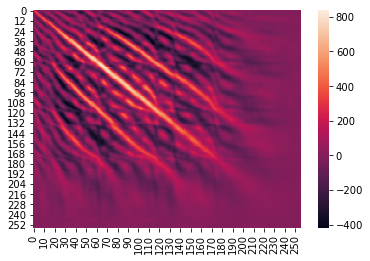}}
\caption{The heatmap shows the dot product similarity of positional embeddings learned by a Transformer model for the En-De WMT32k machine translation task.}
\label{fig:emb}
\end{wrapfigure}

The other category of approaches for PE is to treat each position as a discrete token that can then be uniquely represented as a learnable embedding vector~\cite{visiontransformer,pmlr-v70-gehring17a,kitaev2020reformer,devlin-etal-2019-bert}. The approach can capture arbitrarily complex relationships between positions by learning from data, but it can be difficult to generalize for positions that are rarely encountered during training. For example, the heatmap map in Figure~\ref{fig:emb} shows the positional similarity learned by a Transformer model for a machine translation task on En-De WMT32k~\citep{DBLP:journals/corr/VaswaniSPUJGKP17}. Towards the diagonal, i.e., positions that are closer, there tends to be higher similarity because each token attends to itself the most. However, the trend is diffused for large positions, i.e., when a sequence is long, because fewer training examples have long sequences. For what is followed, a model will not be able to correctly represent large positions in a long sequence at training and test time.

There has been extensive work in extending positional encoding for different modeling tasks, e.g., handling long-range sequences~\cite{dai-etal-2019-transformer,NEURIPS2019_dc6a7e65} or tree structures~\cite{shiv2019novel,wang-etal-2019-self}, or enhancing vision tasks using input-dependent positional encoding~\cite{DBLP:journals/corr/abs-2102-10882}. Our work is related to the effort of using a continuous function instead of embedding retrieval for modeling positions. Previous work~\cite{DBLP:journals/corr/abs-1912-12333} uses complex embedding functions to model 1D positions. It has been shown that position encoding in a 1D space can be learned as a Neural ODE system~\cite{Liu2020LearningTE}. However, their approach cannot be extended to 2D or higher-dimensional problems. More recently, previous work has proposed learnable sinusoidal representations for 1D positions~\cite{wang2021on} in language tasks. In contrast, we focus on representing 2D or even higher dimensional positions in spatial tasks.

%In this work, we focus on encoding multi-dimensional spatial positions using learnable Fourier features to bring in inductive bias. 

Our work is different from the body of work on relative positional encoding, which directly represents pairwise positional relation between query and key~\cite{shaw-etal-2018-self,DBLP:journals/corr/abs-1809-04281,Bello_2019_ICCV,NEURIPS2019_3416a75f,DBLP:journals/corr/abs-2104-01136}. Because
there are $O(N^2)$ of pairwise relations for $N$ positions, relative positional attention is only feasible for a small range, e.g., within a clip distance or local range, although recent work~\cite{pmlr-v139-liutkus21a} has achieved linear complexity by approximating relative positional encoding. Because relative positional encoding directly participates in the computation of the attention matrix, instead of addressing the representation of individual input items, it cannot be easily plugged into many existing Transformer architectures. In contrast, our method is fully compatible with many Transformer benchmarks models.
In our work, we focus on representing individual multi-dimensional spatial positions such that these representations achieve desirable pairwise relation later during attention computation. 

%, even though these shortcomings do not impact the overall accuracy much as these long sequences are less common.

  % by introducing the inductive bias.

%$p\in{R^{N}}$ is an $N$-dimensional position of an input token and $e_{p}\in{R^{M}}$ is an $M$-dimensional vector representation of the position $p$.

\subsection{Encoding Multi-Dimensional Spatial Positions}
\label{section:multidim}
\begin{figure}[ht]
\centering
\subfigure[Sinusoidal-concatenation encoding.]{%
\includegraphics[height=1.5in]{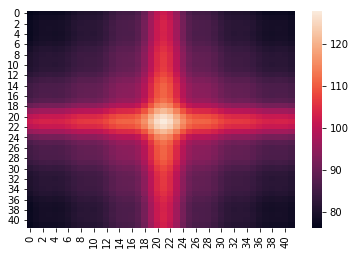}
\label{fig:heatmap_concat}}%
\qquad
\subfigure[2D Fourier feature PE.]{%
\includegraphics[height=1.5in]{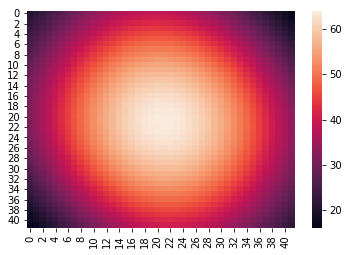}\label{fig:heatmap_fourier}}%
\caption{The similarities of the center position to the rest positions on the 2D space, based on the dot product between their positional encoding of each approach.}
\label{fig:sinusoidal}
\end{figure}

A common approach for positional encoding for a 2D problem is to encode each positional dimension (vertical and horizontal) independently using either sinusoidal (Equation~\ref{eq:sin-cos}) or direct embedding-based methods, and then concatenate these representations to form the final positional encoding~\citep{pmlr-v80-parmar18a,carion2020endtoend,kitaev2020reformer,visiontransformer}. Although the approach of sinusoidal concatenation allows the model to capture the positional (spatial) relationships orthogonally along each axis, the similarity decays much faster along other directions, as shown in Figure~\ref{fig:heatmap_concat}, which ideally should decay at the same rate along all the directions for modeling $L_2$ distances as shown in Figure~\ref{fig:heatmap_fourier}.

%\begin{equation} 
%\label{eq:attention}
%\begin{split}
%\mbox{Attention}(Q,K,V)&=\mbox{softmax}(\frac{QK}{\sqrt{D}})V \\
%Q&=E_{x}M_{Q}\\
%K&=E_{x}M_{K}\\
%V&=E_{x}M_{V}
%\end{split}
%\end{equation}

While concatenating learned embedding has the capacity to model complex spatial relations between positions, they can be difficult to generalize. It is even brittle for addressing problems involving higher-dimensional positions. For example, for modeling spatial structures in UIs~\citep{li-etal-2020-mapping,li-etal-2020-widget}, recent work takes a collection of UI objects as input and the positional attribute of each object is its spatial configuration on the screen, which involves 4 coordinate values: \texttt{[top, left, bottom, right]}. % Although L2 distances are appropriate for the distance between two points, e.g., pixels in images, there are many potential distance metrics between bounding boxes (rectangles) that can be desirable, depending on problems. There can be complex interaction between positional dimensions, which might be difficult to manually specify.
The occurrence of unique object positions can be sparse, which makes it difficult for a model to generalize. There are positions that are rarely seen during training might occur at test time. Motivated by these analyses, we intend to develop a positional encoding method for representing a multi-dimensional position by taking into account all the dimensions holistically, and meanwhile enabling effective inductive bias and learnability in the representation.

\subsection{Fourier Features}
\label{section:fourier}
The task of mapping data points to a vector space such as their dot product achieves certain distance metric has been extensively investigated in the literature of kernel functions~\cite{RFF07,RahimiR08,Quoc14,Yang14,Hamid14,Tewari18}. 
\begin{equation*}
    k(x,y) \approx z(x)'z(y)
\end{equation*}
where $x,y \in \mathcal{R}^d$ and $k(x,y)$ is a shift-invariant kernel function; and $z(x)$ and $z(y)$ are feature mapping respectively. 

Fourier features~\cite{RFF07,RahimiR08} are a common technique to approximate a Gaussian kernel, a shift-invariant kernel, with $k(x,y) = \exp(-\frac{\|x-y\|^2}{\gamma^2})$ where $\|x-y\|^2$ is the Euclidean distance between two points, $x$ and $y$, which each point is a multi-dimensional position in our context. This unique attribute inspired us to represent a multi-dimensional position via Fourier features, which is a basis for our approach for positional encoding.

%Several follow-up works have generalized RFF to other types of kernels, such as polynomial kernels \cite{Felix2015} and inner product kernels \cite{Kar2012RandomFM}. In this work, we focus on Guassian kernel as it closely models L2 Euclidean distances that are desired in popular problems, such as image tasks. 

%There are further research on even speeding up the RFF feature mapping. Fastfood~\cite{Quoc14} sped up the mapping computation by the fast Hadamard transform. 
%Extensive works have been conducted for random features~\cite{Quoc14,Yang14,Hamid14}. 
%\cite{Hamid14} proposes the condensed random feature map to improve performance of RFF. 
%The benefit of random Fourier feature mapping is that it is data independent and computation and memory efficient (linear time/memory complexity w.r.t. the dimension of the data). 
%Due to these merits, 
%Random Fourier feature has been widely used in large-scale machine learning tasks, such as speeding up kernel SVM \cite{Tewari18}. 
Random Fourier features have also been applied in deep learning models, e.g., approximating the attention matrix in Transformer \cite{Performer}. Recently, adaptive random Fourier features~\cite{Li_Zhang_Wang_Kumar_2019} have been proposed for better kernel approximation that show improvement on classification tasks. In contrast, we propose learnable Fourier features for spatial positional encoding and integrate the method in various Transformer-based deep architectures that show improvements on multi-dimensional spatial tasks.

\section{Learnable Fourier Features Positional Encoding}
\label{section:model}

%We first discuss how we design our positional encoder based on random features \citep{RFF07} for modeling L2 distances. We then discuss how to use this encoder as a building block for modeling more complex positional relationships.

%which can represent multi-dimensional positions as vectors whose dot product approximates L2 distances, as discussed in the previous section. 

We propose to learn a position encoding function that maps  an $M$-dimensional position $x\in \mathcal{R}^M$ into a $K$-dimensional feature vector. This $K$-dimensional vector will then be used in downstream computation for attention mechanisms. The proposed encoding function is composed with the following two components: 

\paragraph{Learnable Fourier Features}
To extract useful features from the input position $x$, 
we consider the following feature extraction layer motivated by the idea of Fourier features~\cite{RFF07,RahimiR08}. %Existing sinusoidal position considers the 1D case and encodes $x$ by $\sin (w_ix)$ with a predefined set of wavelength $\{w_i\}$. However, in a higher dimensional case $w$ should be an $M$-dimensional vector 
Given an $M$-dimensional position, $x\in \mathcal{R}^{M}$, we acquire a $D$-dimensional Fourier feature vector representation for the position, $r_{x}\in \mathcal{R}^{D}$, as follows:
\begin{equation}
\label{eq:random_kernel}
r_{x}=\frac{1}{\sqrt{D}}[\cos{xW_{r}^{T}}\mathbin\Vert \sin{xW_{r}^{T}}]    
\end{equation}
where $\mathbin\Vert$ is the concatenation of two vectors. This can also be viewed as the generalization of sinusoidal position encoding to the multi-dimensional case, while we set $W_r\in \mathcal{R}^{\frac{D}{2}\times M}$, which defines both the orientation and wavelength of Fourier features, as trainable parameters. Since $\cos(a-b)=\cos a \cos b + \sin a \sin b$, we have the following:
\begin{equation}
\label{eq:relative}
    r_x \cdot r_y = \frac{1}{D} \text{sum} \big( {\cos((x-y)W_r^T)}\big) := h_{W_r}(x-y)  
\end{equation}
where $\cdot$ is the dot product. Therefore, 
vectors in the form of \eqref{eq:random_kernel} enjoys the shift-invariance property---the dot product of $r_x$ and $r_y$ is a function of $x-y$ and the function is parameterized by $W_r$. Learning $W_r$ is equivalent to obtaining the most informative function on $x-y$ that can be useful for the downstream task. 

%The linear projection, $W_{r}\in R^{\frac{D}{2}\times M}$, 
In our algorithm, the linear projection $W_r$ is initialized by drawing from a normal distribution %with mean 0 and variance $\gamma^{-2}$.
\begin{equation}
\label{eq:guassian}
    W_{r} \sim \mathcal{N}(0,\,\gamma^{-2}). 
\end{equation}
When the linear projection weights are drawn in such a way, according to random Fourier features~\citep{RFF07,RahimiR08}, the dot product between two feature vectors, $r_{x}$ and $r_{y}$, approximates the Gaussian kernel over the original positions.
\begin{equation}
\label{eq:approximate}
    r_{x} \cdot r_{y}\approx\exp(-\frac{\|x-y\|^2}{\gamma^2}).
\end{equation} 
Figure~\ref{fig:heatmap_fourier} visualizes this representation, which introduces a useful inductive bias of $L_2$ distances into the model. 
%\begin{figure}[ht]
%\vskip 0.2in
%\centering
%\centerline{\includegraphics[width=0.8\columnwidth]{figures/random.png}}
%\caption{The similarities of the center position to the rest positions, based on the dot product of random features: $r_{x} \cdot r_{y}$.}
%\label{fig:random}
%\vspace{-2pt}
%\end{figure}

%In our design, each multi-dimensional position is first linearly projected via trainable weights that are initialized based on the random kernel formulation, and then processed by a multi-layer perceptron to generate the final positional encoding. 

%\subsection{Natural Positional Similarity}
%For the one dimensional case, e.g., tokens in a sequence, $s_{ij}^{'}=e^{-|i-j|}$. For the two dimensional case, e.g., individual pixels in an image, $s_{ij}^{'}=e^{-\sqrt{(i_{r}-j_{r})^{2} +(i_{c}-j_{c})^{2}}}$. For the four dimensional case, where the positions in question are for two bounding boxes. We can represent their similarity based on Intersection Over Union. Where each position has four coordinate values: top, left, bottom and right, which we represent as $i_{t}$, $i_{l}$, $i_{b}$, and $i_{r}$.

%\begin{equation}
%    \phi(x) = \frac{1}{\sqrt{N}}[cos(w_1x),\cdots,cos(w_Nx),\\sin(w_1x),\cdots,sin(w_Nx) ]
%\end{equation}
%where $w_i$ is drawn from normal distribution with mean 0 and variance $\gamma^{-2}$.

\paragraph{MLP layer}
To feed the representation to the downstream computation, we give the representation additional capacity by modulating the features with a multi-layer perceptron: 
\begin{equation}
    \label{eq:mlp}
    PE_{x}=\phi(r_{x},\theta)W_{p}, 
\end{equation}
where $\phi(\cdot)$ is the perceptron parameterized by $\theta$. $W_{p}$ are trainable parameters for projecting the representation onto a target dimension of positional encoding for combining with content embedding. Our purpose with MLP here is very different from previous work that uses non-linear transformation such as an RNN to capture positional dynamics~\cite{neishi-yoshinaga-2019-relation,Liu2020LearningTE}. These previous works do not handle non-sequential multi-dimensional positions.

The learnable parameters in our position encoding function are $W_r$ for Fourier features and $\theta, W_p$ for the MLP layer. The size of these matrices are independent of the sequence length. Furthermore, the position encoding function can be applied to any input position $x$, so our method can be easily applied when training and testing involve different positions, e.g., images with different resolutions. Compared to the previous sinusoidal representation (Equation~\ref{eq:sin-cos}), our representation is learnable and multi-dimensional. Compared to the discrete embedding-based approach, our representation treats each dimension of a position as a continuous-valued vector, which alleviates the sparsity issue with discrete positions. Previous work has revealed that using sinusoidal activation functions might suffer optimization problems due to vanishing gradients in extreme cases~\cite{Parascandolo2017TamingTW}, although we do not observe much difficulty in training our  positional encodings.

%It is seeded by drawing from a Gaussian distribution to bring in the inductive bias of L2 distances. To approximate L2 distances with dot product of their vector representation (Equation~\ref{eq:approximate}), $W_{r}$ needs to obey a Gaussian distribution centered at $0$ (see Equation~\ref{eq:guassian}). Although $W_{r}$ are seeded by random weights sampled from Gaussian, it is trainable, which can change during learning. %Thus, we introduce a regularizer loss based on the KL divergence between the distribution of $W_{r}$ and a desired Gaussian distribution.

%kl_loss = -0.5 * (
%        1 - jnp.log(ideal_var) + jnp.log(var) - (var + mu * mu) / ideal_var)
%\begin{equation}
%    \Lagr_{KL}=-\frac{1}{2}(1-\log\bar{\sigma}^{2}+\log\sigma^{2}-\frac{\sigma^{2}+\mu^{2}}{\bar{\sigma}^{2}})
%\end{equation}

%where $\mu$ and $\sigma^{2}$ are the mean and variance of $W_{r}$. $\bar{\sigma}^{2}$ is the target variance that is %also learnable, which is initialized as 
%$\gamma^{-2}$. When training the model, the regularizer loss $\Lagr_{KL}$ is added to the overall loss for optimization.
%\vspace{-3pt}
%\begin{equation}
%\label{eq:regularizer}
%    \Lagr_{total}=\Lagr_{model}+\alpha \Lagr_{KL}
%\end{equation}
%\vspace{-1pt}
Our representation is applicable for many 2D spatial tasks, e.g., image-related tasks. 
For tasks involving higher-dimensional positions, the positional similarity between positions might be more complicated than $L_2$ distances. For example, to model the spatial structure of a natural scene or a graphical user interface, given two objects in the structure, $x$ and $y$, coordinate values, $[x_{1}, x_{2}, x_{3}, x_{4}]$ and $[y_{1}, y_{2}, y_{3}, y_{4}]$, represent the object's top, left, bottom, and right position. The $L_2$ distance between the two positions $\sum_{i=1}^{4}(x_{i}-y_{i})^{2}$ will capture neither the minimum nor the maximum distance between the two objects, or any vertical or horizontal alignments of them. To address this issue, we hypothesize that complex spatial relationships can be built on top of shift-invariant relations enabled by our positional encoding. Specifically, we can partition a multi-dimensional position into groups, and apply the same encoding pipeline to each group of coordinate values. The process is similar to applying convolution over partitions with the kernel and stride sizes to be the group size. We can then concatenate the output of all the groups to form the final positional encoding. We will elaborate on this use case in the UI modeling experiment (Section~\ref{sec:widget_captioning}).
%For the UI modeling example, instead of passing the four coordinate values of a bounding box through the positional encoder, we can pass the \texttt{(top, left)} and \texttt{(bottom, right)} through the encoder separately,%, or pass in \texttt{(top)}, \texttt{(left)}, \texttt{(bottom)}, and \texttt{(right)} all separately, 
%although they all share parameters. The downstream linear projection of keys and queries gives the model capacity to mix and match these components if needed. 
An implementation of our positional encoder based on tensor operation is detailed in Algorithm~\ref{alg:psedo1} in which Equation~\ref{eq:random_kernel} and~\ref{eq:mlp} are realized in Line 1 and 2. 

%Given a $d$-dimensional position vector say $x = (x_1, x_2, \cdots, x_d)$, if it is an image, then $d=2$; if it is a video, $d = 3$; etc, we want to have the position encoding for $x$.

%\begin{equation}
%    \phi(x) = \frac{1}{\sqrt{N}}[cos(w_1x),cos(w_2x),\cdots,cos(w_Nx),sin(w_1x),sin(w_2x),\cdots,sin(w_Nx) ]
%\end{equation}

%By such mapping, we can make sure given any two positions $x$ and $y$:
%\begin{equation}
%    \phi(x)^T\phi(y) = exp(-\frac{\|x-y\|^2}{2\gamma^2} )
%\end{equation}

%Advantage:
%\begin{itemize}
%    \item No assumption about different sizes of the image, no matter it is 28*28 size or 64*64.
%    \item It can generalize to new positions without any other calculation
%    \item Keep the desired distance
%    \item Deterministic---as long as we keep the same random seed (YL: Assume we only need to calculate the weights once).
%    \item close positions will have small distance with distance exponentially decreases. (YL: Gaussian or RBF kernel is actually proportional to attentional aligments due to softmax).
%\end{itemize}

%where $e_i\in R^{|e|}$ and $|e|$ is the depth of the embedding. $\phi(\cdot)$ is the embedding function that can be as simple as just embedding lookup or it can be a multi-layer perceptron parameterized by $\theta$. 

\begin{algorithm}[ht]
\caption{Compute the Fourier feature positional encoding of a multi-dimensional position.}
\label{alg:psedo1}
 \KwIn{A tensor $X$ in the shape of $[N, G, M]$ that represents $N$ positions where each position is in the shape of $[G, M]$ that represents $G$ positional groups and each group has $M$-dimensional positional values.}
 \KwOut{$PE_{X}$ in the shape of $[N, D]$ where $D$ is the depth of the positional encoding.}
 
{\nonl{\bf{Hyperparameter}}: The depth of the Fourier feature dimension $|F|$, the hidden layer dimension $|H|$, and the positional encoding dimension $D$, and $\gamma$.}

{\nonl {\bf{Initialization}}: Initialize learnable weights $W_{r}\in \mathcal{R}^{\frac{|F|}{2}\times M}$ by sampling from $\mathcal{N}(0,\ \gamma^{-2})$; Initialize learnable weights $W_{1}\in \mathcal{R}^{|F|\times |H|}$, $B_{1}\in \mathcal{R}^{|H|}$, $W_{2}\in \mathcal{R}^{|H|\times \frac{D}{G}}$ and $B_{2}\in \mathcal{R}^{\frac{D}{G}}$.}

{\nonl \ }

$F\leftarrow\frac{1}{\sqrt{|F|}}[\cos{XW_{r}^{T}}; \sin{XW_{r}^{T}}]$ (Eq.~\ref{eq:random_kernel})\; 

$Y\leftarrow\mbox{GeLU}(FW_{1}+B_{1})W_{2} + B_{2}$ (Eq.~\ref{eq:mlp}) \;

%$Y=\frac{x}{2} \cdot (1 + \hbox{erf}(\frac{x}{\sqrt{2}}))$

$PE_{X}\leftarrow$ Reshape $Y$ into the shape of $[N, D]$\;

%$\mu \leftarrow$ $\mbox{Average}(W_{r})$\;

%$\sigma^{2} \leftarrow$ $\mbox{Variance}(W_{r})$\;

%$\Lagr_{KL} \leftarrow$ $-\frac{1}{2}(1-\log\bar{\sigma}^{2}+\log\sigma^{2}-\frac{\sigma^{2}+\mu^{2}}{\bar{\sigma}^{2}})$

%{\nonl \;}

 %\Return $P_{X}$ and $\Lagr_{KL}$.
 
 \Return $PE_{X}$.
\end{algorithm}

\section{Experiments}
We evaluate our approach on a range of benchmark tasks using Transformer-based models in comparison with several existing positional encoding methods. %These tasks include image generation with Reformer \citep{kitaev2020reformer}, object detection with DETR \citep{carion2020endtoend}, image classification with Vision Transformer \citep{visiontransformer}, and natural language generation in graphical user interfaces \citep{li-etal-2020-mapping}.

\subsection{Image Generation}
We compare our method with existing positional encoding approaches based on Reformer~\citep{kitaev2020reformer} for the image generation task on the ImageNet 64x64 dataset~\citep{DBLP:journals/corr/abs-1904-10509}. Reformer is a Transformer-based model that uses locality-sensitive hashing and reversible residual layers to efficiently handle long sequences. % where attention would be otherwise very expensive to calculate. 
Reformer flattens a 64x64 image into a sequence (Length=64x64x3=12,288) in a raster scan red-green-blue order. Reformer as an auto-regressive model predicts the pixel value at each position by attending to previous positions. We equip Reformer with different positional encoding methods. 
\begin{itemize}
    \item \textit{Embed-2D}: Reformer's default positional encoding concatenates the embedding of each dimension from two embedding matrices: vertical $[64, 384]$ and horizontal $[64, 384]$.
    \item \textit{Embed-1D}: The baseline method assigns a learnable embedding to each position in the flattened sequence, from an embedding matrix of $[64\times 64, 768]$, which ignores the 2D structure of an image and lets the model learn positional relations all by itself.
    \item \textit{Sine-2D} and \textit{Sine-1D}: Similar to Embed-2D and Embed-1D, but they instead encode a position using Transformer's constant sinusoidal formulation (Equation~\ref{eq:sin-cos}). %One form is to fix the representation and the other form is to use these representations to initialize trainable embedding weights.
    \item \textit{Learnable-Fourier + MLP}: Our method that implements Algorithm~\ref{alg:psedo1} using the hyperparameter $|F|=384$, $|H|=32$, $D=768$. We picked these dimensions for our method to have roughly the same number of parameters as Embed-2D, the benchmark of Reformer.
\end{itemize} 
% because vertical and horizontal positions need to be mapped jointly to model L2 distances on an image. 
We leave the RGB axis to use the direct embedding as the original Reformer: $[3, 256]$. The concatenation of the pixel position encoding and the RGB index embedding results in an representation that has the same depth ($1024$) as the one in the original paper, which allows the rest of the model intact. 

We follow the experimental procedure as detailed in the Reformer paper. All our experiments used a 6-layer, 8-head-attention Reformer, with $d_{model} = 1024$, $d_{ff} = 4096$, and $n_{heads} = 8$. These models are implemented based on the Reformer codebase in Trax\footnote{\url{https://github.com/google/trax/tree/master/trax/models/reformer}}. The training for each Reformer model is parallelized across 32 TPU v2 cores, and each batch contains 8 sequences (images) on each core. We trained each model variant for 100k steps, which took about 24 hours to complete.

\begin{figure}[t]
\centering
\subfigure[Comparison w/ baselines.]{%
\includegraphics[height=1.65in]{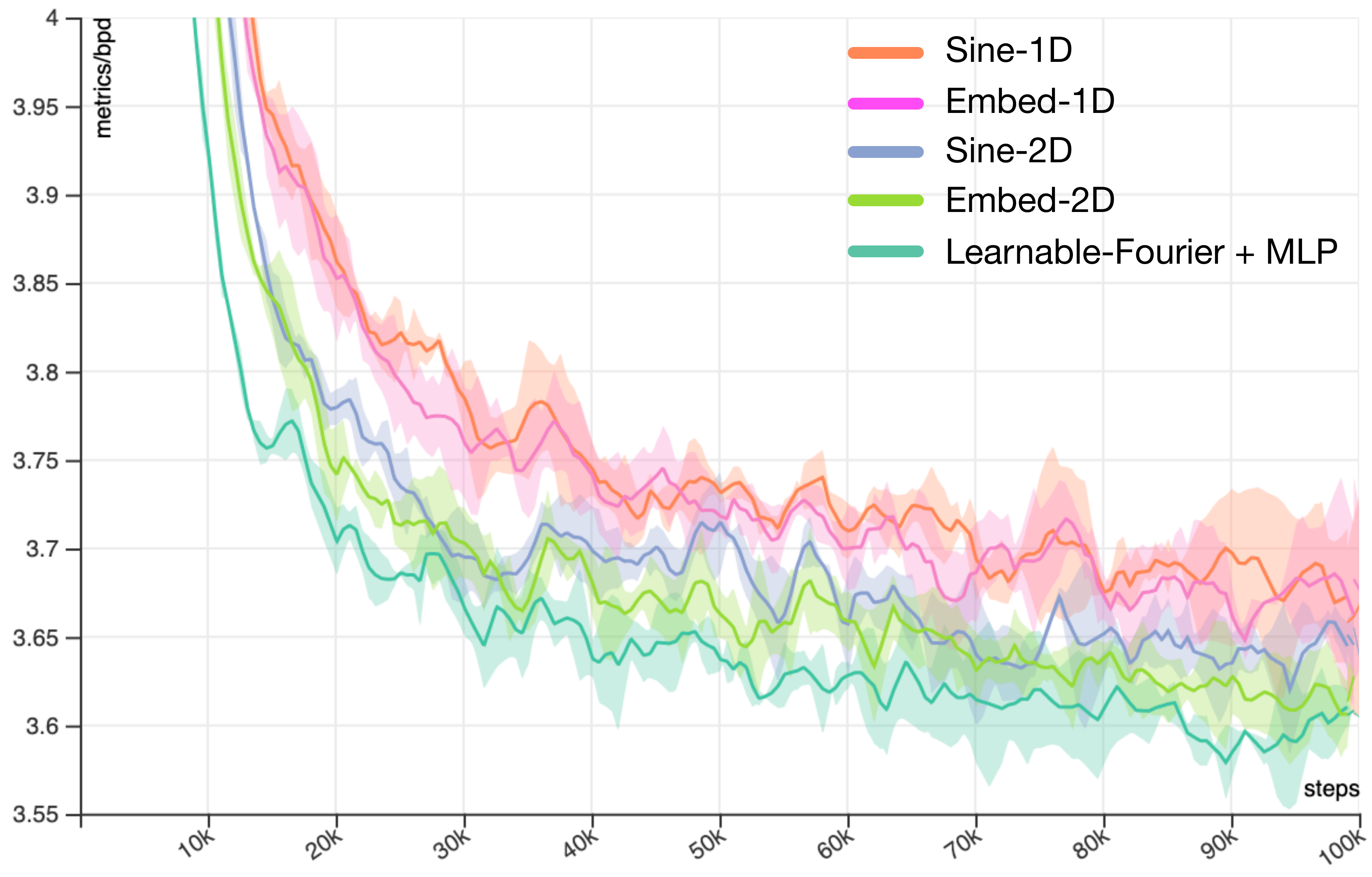}
\label{fig:reformer_baseline}}%
\qquad
\subfigure[Comparison w/ individual components.]{%
\includegraphics[height=1.65in]{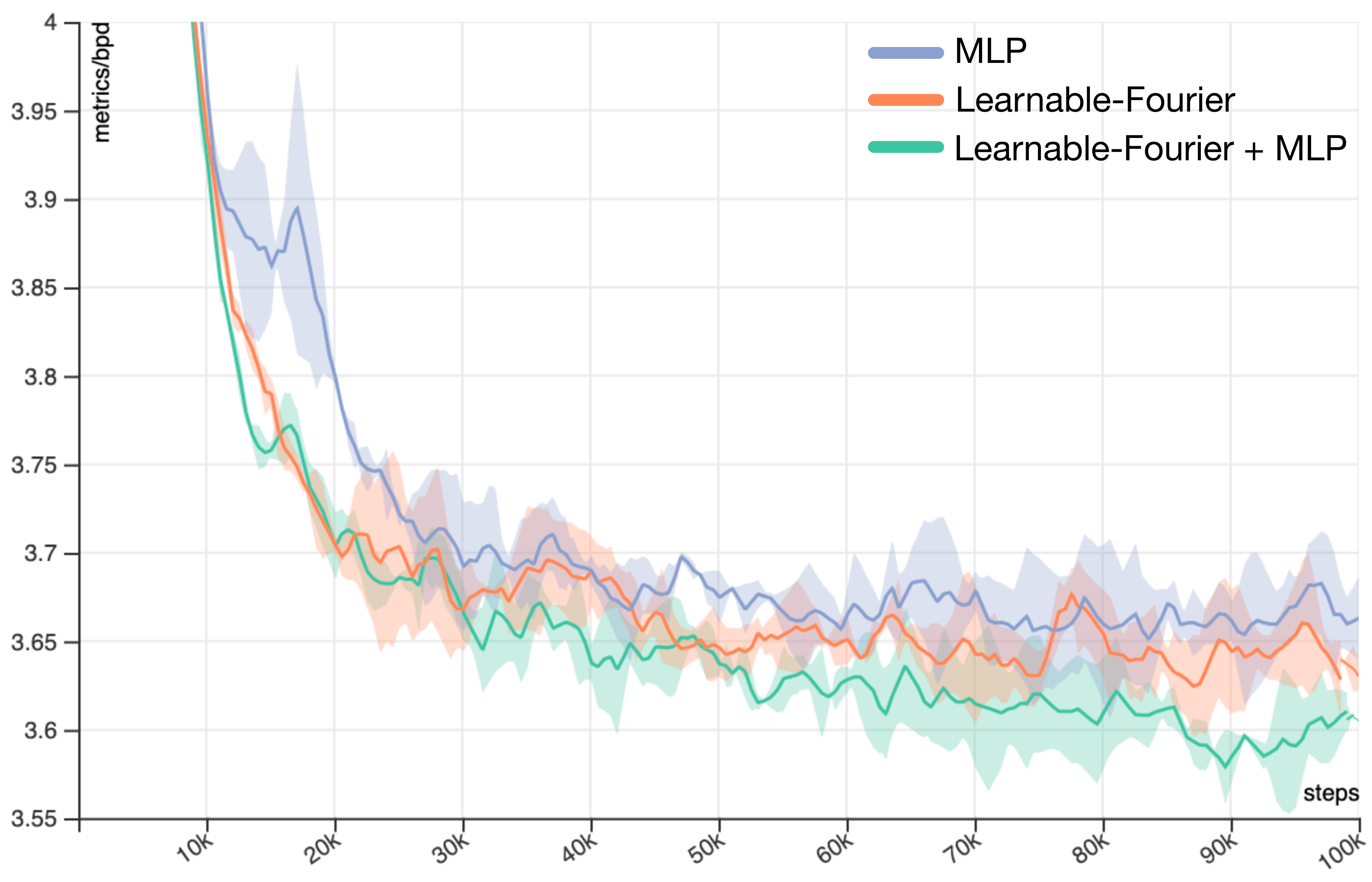}\label{fig:reformer_transformer}}%
\qquad
%\subfigure[Comparison w/ MLP.]{%
%\includegraphics[height=1.7in]{figures/reformer-mlp.pdf}\label{fig:reformer_mlp}}%
%\qquad
%\subfigure[Fourier ablation.]{%
%\includegraphics[height=1.7in]{figures/reformer-fourier.pdf}\label{fig:reformer_fourier}}%
\caption{Bits per dim (bpd) w.r.t. training steps on evaluating Reformer on the held-out data of the ImageNet 64x64 dataset for image generation, using different positional encoding methods. The plot shows the mean and 95\% confidence interval based on 3 repeats of experiments for each method.}
\label{fig:reformer_imagenet64}
\end{figure}

As shown in Figure~\ref{fig:reformer_imagenet64}a, our method, Learnable-Fourier + MLP, outperforms all the baselines in terms of convergence speed and achieves better accuracy, i.e., lower bits per dim at the end. The Reformer's original positional encoder, Embed-2D, is the second best. Sine-2D clearly outperforms Sine-1D, and Embed-1D achieves a similar performance as Sine-1D. 

To understand how each component in our method contributes to the overall performance, we compare Learnable-Fourier+MLP with its components Learnable-Fourier and MLP alone. MLP takes a 2D position as input and outputs a 768-dimensional positional encoding. Our experiment shows that Learnable-Fourier or MLP alone does not perform as good as their combination, Learnable-Fourier+MLP (see Figure~\ref{fig:reformer_imagenet64}b). It is worth noting that Learnable-Fourier shows competitive performance for the first 30k steps, which indicates that it benefits from an effective bias for capturing meaningful positional relationships. %See Appendix~\ref{sec:ablation} for additional ablation studies.

%Embedding 1D: /cns/oz-d/home/liyang/rs=6.3/reformer_v14

%Embedding 2D: /cns/oz-d/home/liyang/rs=6.3/reformer_axial_v14

%Sinusoidal 1D: /cns/oz-d/home/liyang/rs=6.3/reformer_sin-cos_v14

%Sinusoidal 2D: /cns/oz-d/home/liyang/rs=6.3/reformer_sin-cos2_v21_oz

%Learnable RF: /cns/oz-d/home/liyang/rs=6.3/reformer_random_small_bias_v22_oz

%Embedding 2D (Axial): Total number of trainable weights: 60664832
%https://xm2a.corp.google.com/experiments/20040007/
%Learnable RF: #params=60668385 /cns/oz-d/home/liyang/rs=6.3/reformer_random_small_bias_v22_oz
%Sinusoidal: #params: 60614912 https://xm2a.corp.google.com/experiments/20062598
%Embedding 1D: #params=73197824

\subsection{Object Detection}
\label{section:detr}
We evaluate the proposed positional encoding in DETR~\citep{carion2020endtoend}, a recent model that uses a Transformer for end-to-end object detection. It uses a Transformer to take the output from a ResNet, i.e., a feature map with the spatial dimensions of $42\times42$. Similar to Reformer, positional encoding represents each position in the grid as part of the input to the Transformer encoder in DETR. 
We experiment with the default 6-layer Encoder-Decoder setup in DETR, with the same set of hyperparameters, on the COCO 2017 object detection dataset~\cite{DBLP:conf/eccv/LinMBHPRDZ14} that has 118k images for training and 5k for validation. 
We equip the DETR model with different positional encoding methods, including Sine-2D that is DETR's default method, Learnable-Fourier+MLP, Embed-2D and MLP. The model implementations are based on the DETR codebase\footnote{\url{https://github.com/facebookresearch/detr/blob/master/models}}, which are ported into JAX\footnote{\url{https://github.com/google/jax}}. The training for each DETR model is parallelized across 64 TPU v3 cores with a batch size of 64 images. We let each model train for 300 epochs to converge, which took about 3 days. We follow the experimental procedure of the DETR paper, and report accuracy on the validation set. 

%In this experiment, we focus on how different positional encoding methods impact the convergence of learning.

%\begin{figure}%
%\centering
%\includegraphics[height=1.9in]{figures/detr2.pdf}
%\label{fig:detr}%
%\caption{The impact of different positional encoding methods on the DETR model, shown as $AP$, $AP_{75}$ from left to right, on validation data as training progresses for the initial 100k steps.}
%\end{figure}

DETR uses image augmentation in both training and validation. Each image is randomly resized to several specific dimensions with the smaller side of the image at one of the following sizes: 480, 512, 544, 576, 608, 640, 672, 704, 736, 768, and 800. For positional encoding, all image positions are normalized to a range of $(0, 1)$. Normalization is valuable because of random resizing and cropping during image augmentation results in images with different sizes. Embed-2D treats each position as a discrete value, and all the methods except Embed-2D leverages position normalization. As shown in Table~\ref{tab:obj_normal}, Learnable-Fourier+MLP offers the best performance across all the metrics. %We see a bigger gain of our method on the $AP_{75}$ metric that requires more precise matches between ground-truth and predicted bounding boxes. 
Sine-2D and MLP perform competitively while Embed-2D has the worst performance. 

To investigate how each encoding generalizes to unseen image sizes, we modify the benchmark by reserving the three largest sizes: 736, 768, and 800 for validation only. We also disable position normalization. As a result, there are large positions that are never seen during training, which requires each method to generalize (or extrapolate) to these positions. As shown in Table~\ref{tab:obj_extra}, the benefit of Learnable-Fourier+MLP is more pronounced, and the performance gap between Embed-2D and the other methods is further increased.

\begin{table}
  \caption{The impact of different positional encodings on DETR for object detection.}
  \label{tab:obj_normal}
  \centering
  \begin{tabular}{lllllll}
    \toprule
    Method     & $AP$     & $AP_{50}$ & $AP_{75}$     & $AP_{small}$ & $AP_{medium}$ & $AP_{large}$\\
    \midrule
    Sine-2D     & 40.1 & 60.4 & 42.6 & 18.5 & 43.6 & 58.8      \\
    Embed-2D     & 39.3 & 59.8 & 41.4 & 18.7 & 42.5       & 57.5  \\
    MLP     & 40.0       & 60.3 & 42.2 & 18.6 & 43.7 & 58.1  \\
    Learnable-Fourier+MLP & \bf 40.2  & \bf 60.7 & \bf 42.7 & \bf 18.8 & \bf 43.8 & \bf 59.1     \\
    \bottomrule
  \end{tabular}
\end{table}

\begin{table}
  \caption{Performance of each method for object detection involving unseen image dimensions.}
  \label{tab:obj_extra}
  \centering
  \begin{tabular}{lllllll}
    \toprule
    Method     & $AP$     & $AP_{50}$ & $AP_{75}$     & $AP_{small}$ & $AP_{medium}$ & $AP_{large}$\\
    \midrule
    Sine-2D     & 38.9 & 59.6 & 40.9 & 17.5 & 42.5 & 57.5      \\
    Embed-2D     & 36.6 & 58.2 & 37.7 & 15.9 & 40.0       & 55.3  \\
    MLP     & 38.6       & 59.5 & 40.3 & 17.1 & 42.1 & 57.1  \\
    Learnable-Fourier+MLP & \bf 39.5  & \bf 60.0 & \bf 41.6 & \bf 18.9 & \bf 43.0 & \bf 58.0     \\
    \bottomrule
  \end{tabular}
\end{table}

%The difference between methods is small with such a large model, positional normalization and extensive training, although the proposed PE is still consistently better. The performance are generally better than the extrapolation case shown above.

%Regarding Figure 4, we trained each model for a fixed number of steps to show the convergence rate for the initial 90k steps. To answer your question, we conducted additional experiments to train these models fully (300 epochs=554.4k steps) to convergence. Here are their final performance. The difference between methods is small with such a large model, positional normalization and extensive training, although the proposed PE is still consistently better. The performance are generally better than the extrapolation case shown above.

\subsection{Image Classification}
We evaluate the proposed positional encoding on image classification, another popular task on images, based on Vision Transformer (ViT)~\cite{visiontransformer}, a Transformer-only architecture that does not use CNN for image embedding. The default positional encoding in ViT is Embed-1D. In this experiment, we focus on the ViT-B/16 model that is a 12-layer Transformer encoder with Hidden\_size=768, MLP\_size=3072 and 12 attention heads.
The input to ViT-B/16 uses a $14\times 14$ image grid where each cell corresponds to a $16\times 16$ image patch. We train each model on the ImageNet dataset for 90 epochs, and report its accuracy on the ImageNet validation dataset. Learnable Fourier+MLP achieved better performance (Precision@1=74.5\%) on the validation dataset than Embed1D (Precision@1=73.6\%). 

Dosovitskiy et.al. investigated several positional encoding methods in their work (see Table 8 in~\cite{visiontransformer}), including Embed-1D, Embed-2D and relative positional encoding. They pre-trained these models on the large JFT dataset (300 million examples) and then report their performance on ImageNet 5-shot linear tasks. They found the model suffers when no positional encoding is used but there are no significant impacts for using each of these positional encoding methods. We suspect that given such a large model (86M Params), it is not difficult for any of these positional encoding methods to learn the small number of unique positions ($14\times 14=196$) on the image. In their experiment, Embed-1D achieves 64.206\% accuracy, Embed-2D 64.001\% and Relative Positional Encoding 64.032\%. We experimented Learnable-Fourier+MLP in this experiment, which achieved 64.732\% accuracy.

\subsection{Widget Captioning}
\label{sec:widget_captioning}

So far, we have investigated tasks that handle 2D positions in an image. In this experiment, we investigate even higher-dimensional positions. %Recently, Transformer-based models have been used for encoding user interface screens~\citep{li-etal-2020-mapping,li-etal-2020-widget}. %, which come with a collection of objects and each object are present on the screen as a rectangular region. 
In a widget captioning task~\citep{li-etal-2020-widget}, the model is trained to generate natural language description of widgets in graphical user interfaces, e.g., buttons and icons. A significant part of the model is to encode a UI screen structure, which consists of a collection of 2D objects of different sizes, using a Transformer encoder. To represent the spatial configuration of each object, the original model assigns a learnable embedding vector to every discrete coordinate value of each dimension of the object bounding box, including the left, top, right, and bottom dimensions. The four embedding vectors then jointly represent a bounding box on the screen. We refer this baseline as \textit{Embed-4D}. Li et al. found that position encoding has a significant impact on the performance of widget captioning models (see Table 6 in Appendix F of the paper~\cite{li-etal-2020-widget}).

Because there is no obvious distance metrics between bounding boxes, we hypothesize that an appropriate metric can be learned on top of $L_2$ distances of specific dimensions. To do so, 
We evaluate three different partitions of bounding box dimensions, and use our method to encode each group in parallel as detailed in Algorithm~\ref{alg:psedo1}: Learnable-Fouier+MLP-1/4 treats all the 4 coordinate dimensions \texttt{[(top, left, bottom, right)]} as one group, i.e., $G=1$; Learnable-Fourier+MLP-2/2 splits the 4 dimensions into 2 groups \texttt{[(top, left), (bottom, right)]}, i.e., $G=2$; and finally Learnable-Fourier+MLP-4/1 encodes 4 groups of 1-dimensional value \texttt{[(top), (left), (bottom), (right)]}, i.e., $G=4$.
%we encode the four coordinates separately into a vector of size 32, and concatenate them as the final position embedding.
 We also add the sinusoidal approach to the comparison, which represents each positional dimension separately and then uses their concatenation to as the positional encoding a bounding box (referred as \textit{Sine-4D}). 

%We follow the experimental procedure and use the winner model architecture from the paper~\citep{li-etal-2020-widget} based on the public code base\footnote{\url{https://github.com/google-research/google-research/tree/master/widget_caption}}.

\begin{table}[ht!]
\caption{The performance of different positional encoding methods on the widget captioning test set. SOTA shows the results from the original paper, which is reproduced by Embed-4D in our experiment.}
\label{tab:caption_accuracy}
\centering
\begin{tabular}{lcccccc}
    \toprule
    {Method}
    & {BLEU-1}
    & {BLEU-2}
    & {ROUGE}
    & {CIDEr}
    & {METOER}
    & {SPICE} \\
    \midrule
    SOTA~\cite{li-etal-2020-widget} & 44.9 & 32.2 & 44.7 & 97.0 & 31.7 & 17.6  \\
    
    Embed-4D & 45.2 & 31.9 & 45.0 & 97.0 & 31.7 & 17.3 \\
    % avg. scores for t-test
    % Embed-4D & 45.1 & 32.2 & 44.6 & 97.5 & 31.5 & 17.6 \\
    
    MLP & 34.0 & 23.5 & 33.7 & 70.3 & 23.7 & 10.2 \\
    
    Sine-4D & 44.9 & 31.9 & 43.9 & 94.9 & 31.0 & 16.7 \\
    Learnable-Fourier-2/2 & 44.9 & 31.6 & 44.3 & 95.3 & 31.6 & 17.7 \\

    Fixed-Fourier+MLP-1/4 & 45.0 & 32.1 & 44.2 & 95.4 & 31.2 & 17.1 \\
    Fixed-Fourier+MLP-2/2 & 46.1 & 32.5 & 45.8 & 100.2 & 32.5 & 18.4 \\
    Fixed-Fourier+MLP-4/1 & 45.5 & 32.1 & 45.1 & 97.2 & 31.7 & 17.6 \\
    
    Learnable-Fourier+MLP-1/4 &  45.6 & 32.7 & 45.2 & 99.1 & 32.2 & 17.1 \\
    
    Learnable-Fourier+MLP-2/2 & 46.1 & 32.7 & 45.9 & 98.0 & \textbf{32.6} & \textbf{17.9} \\
    % previous last ckpt scores
    % Learnable-Fourier+MLP-4/1 & \textbf{46.1} & \textbf{32.7} & 45.2 & \textbf{99.2} & 31.8 & \textbf{18.1} \\
    % best scores of last ckpt of 3 runs.
    Learnable-Fourier+MLP-4/1 & \textbf{46.8} & \textbf{33.4} & \textbf{46.1} & \textbf{100.7} & 32.4 & 17.8 \\
    
    % avg. scores for t-test
    % Learnable-Fourier+MLP-2/2 & 45.7* & 32.5 & 45.1* & 97.9 & 31.9* & 17.6 \\
    % Learnable-Fourier+MLP-4/1 & \textbf{46.1}* & \textbf{32.9} & \textbf{45.6}* & \textbf{99.3}* & \textbf{32.3}* & \textbf{17.8} \\
   \bottomrule
\end{tabular}
\end{table}

We use the same model architecture and hyperparameters of the strongest model, \textit{Pixel+Local+Context}, as the original paper \citep{li-etal-2020-widget}, and built our experiment based on the public codebase of widget captioning\footnote{\url{https://github.com/google-research/google-research/tree/master/widget_caption}}. Specifically, the screen encoder uses a 6-layer, 8-head Transformer with a hidden size of 128. We train all the models to 100k steps with Adam optimizer and a scheduled learning rate detailed the original paper. %—--a linear warmup followed by an exponential decay.
All the models converged within 12 hours using 4 V100 GPU cores. 
%The number of trainable parameters for the models with Embed-4D, Sine-4D, Learnable-Fourier+MLP-1/4, Learnable-Fourier+MLP-4/1, and Learnable-Fourier+MLP-2/2 are 5.11M, 5.07M, 5.11M, 5.07M and 5.07M, respectively.

All the results are acquired by applying each trained model on the test dataset, based on the same set of captioning metrics. 
As shown in Table~\ref{tab:caption_accuracy}, our method outperforms the benchmark method Embed-4D (\#Params=5.11M) with a large margin even though our method uses fewer parameters (\#Params=5.07M), particularly on BLEU-1, BLEU-2, ROUGE and CIDEr, which clearly advanced the state of the art for this task. Interestingly, both Learnable-Fourier+MLP-2/2 and Learnable-Fourier+MLP-4/1 outperform Learnable-Fourier+MLP-1/4, which indicate that more complex distances needed to be modeled in this task than $L_2$ distances. Compared to Embed-4D, T-tests (over 3 runs of each model) show the gain of Learnable-Fourier+MLP 4/1 is statistically significant ($p < 0.05$) on BLEU-1, ROUGE, CIDEr and METOER; Learnable-Fourier + MLP 2/2 achieves significance ($p < 0.05$) on BLEU-1, ROUGE and METOER. For the two champion conditions, i.e., Learnable-Fourier+MLP-4/1 and 2/2, we found on most metrics there is no statistical significance between their performance  ($p>0.05$). Learnable-Fourier+MLP-4/1 outperforms 2/2 only on CIDEr with marginal statistical significance ($p=0.042$).

We also included a few ablation studies in this experiment. One variant is to fix Fourier features but still include MLP. In this group, i.e., Fixed-Fourier+MLP-*, Fixed-Fourier+MLP-2/2 clearly performs the best across all the metrics. Overall, it seems that Learnable-Fourier+MLP still has advantages over the fixed one on most cases. We then look at Learnable-Fourier but without using MLP. Learnable-Fourier-2/2 seems to perform worse than its counterpart in the other groups on every metric, which indicates that MLP is a crucial component for positional encoding in this task. Lastly, although using MLP alone as the encoding function seems competitive in the object detection task, it performs poorly in this experiment.

%Learnable-Fourier+MLP-2/2 and Learnable-Fourier+MLP-4/1 give the downstream layers in the model the opportunity to learn a desired function using L2 distances from these positional encoder as building blocks.

\section{Discussion}
One clear trend that emerges from our experiments is that positional encoding methods that treat an image as a flattened sequence (Embed-1D or Sine-1D) do not perform well, even though the model is given a great capacity to learn these positional relations. We also observe that taking a multi-dimensional position holistically often performs better than representing each dimension separately and then concatenating these representations. %- Embedding 1D works okay for Vision Transformer but not for Reformer. Because Reformer has a much larger grid size.
%All these benchmark models that we base our experiments on combine positional encoding and content embedding via addition, instead of concatenation (see Equation~\ref{eq:embedding}). 
We found it generally beneficial to use the multi-layer perceptron (Equation~\ref{eq:mlp}) to process the Fourier features for positional encoding before it is mixed with content embedding. We obtained mixed results for using MLP alone as the positional encoding function, which performs competitively on the object detection task but poorly on the UI modeling task that involves sparse spatial structures.  %When concatenation is used as the combiner, the multi-layer perceptron might not be needed because the downstream projection (Equation~\ref{eq:qkv}) will have the capacity to merge the two appropriately. A quick experiment on Reformer using concatenation as the combiner gave us promising results.
From these experiments, it seems not necessary to use a large random feature dimension to achieve good results. %We also found that the standard deviation of the weight distribution evolves during learning, which indicates the scale of relative positional similarity. 

\begin{table}[h]
  \caption{The accuracy of each method on widgets with seen and unseen positions.}
  \label{tab:oov_accuracy}
\centering
\begin{tabular}{lcc}
    \toprule
    {Method}
    & {Seen CIDEr}
    & {Unseen CIDEr} \\
    \midrule
    Embed-4D & \textbf{123.4} & 78.5 \\
    Sine-4D & 121.3 & 76.4 \\
    Learnable-Fourier+MLP-4/1 & \textbf{123.4} & \textbf{82.2} \\
   \bottomrule
\end{tabular}
\end{table}

\begin{figure}
\centering
\includegraphics[width=1.0\textwidth]{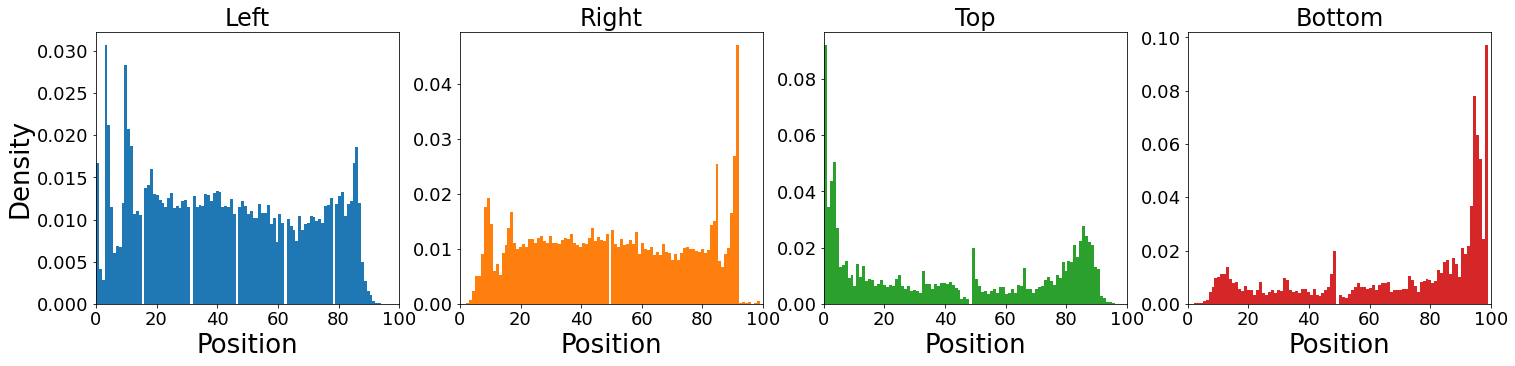}
\caption{Widget position distributions for each dimension in the training set. There are positions rarely occurred in the training set.}
\label{fig:widget-position}
\end{figure}

To understand how different positional encoding methods can generalize to unseen positions, we analyze test results for the widget captioning task. There are positional values rarely or never seen in the training set (Figure \ref{fig:widget-position}). 
%We define a position "Unseen" when its coordinate tuple does not appear in the training data. 
Specifically, 1867 widgets in the test set have seen positions and 2692 have unseen positions. Table \ref{tab:oov_accuracy} shows that our method generalizes to unseen positions significantly better than baselines.
There are a number of reasons for the proposed positional encoding to generalize for unseen positions. First, it treats positions as continuous-valued vectors. As a result, it does not suffer from the difficulty with embedding-based approaches where an embedding vector is assigned to a discrete position, which can be not trained or significantly under-trained when a position is unseen or rarely seen. Second, the Fourier features capture the relative positional relationships by maintaining the shift-invariant property during learning (Equation~\ref{eq:relative}), which applies to unseen positions as well.

One direction that deserves further investigation is how the interaction between positional encoding and content embedding should be taken into account for the design of a positional encoding function. Our work investigated positional encoding when it is combined with content embedding via addition for all the image tasks. It would be interesting to investigate how our positional encoding performs when it is concatenated with content embedding on these tasks. 

Although Euclidean distances might be a desirable positional metric for images, tasks such as widget captioning involves sparse spatial structures, and the spatial relations between two rectangular objects on the screen can be more complicated. For example, similarity between two bounding boxes can be related to their vertical or horizontal alignment or overlaps (IoU), or other domain specific factors related to UI layouts. Our positional encoding method outperformed benchmark methods in this task, which showed that it is better equipped to capture these spatial relationships. Yet, it is worth investigating methods that can more directly capture these complex spatial relationships.

\section{Conclusion}
We present a novel approach for positional encoding based on learnable Fourier features. %It naturally approximates L2 distances that are desired in many image-based tasks and can be used as a building block to address problems with even higher-dimensional positions. 
We evaluate our approach on a range of multi-dimensional spatial tasks, including image generation, object detection, image classification, and sparse spatial structure modeling in user interfaces, which show that our positional encoding consistently outperforms the benchmark methods.

\section*{Acknowledgments}
We would like to thank anonymous reviewers for their insightful comments and constructive feedback that have significantly improved the work.

\bibliographystyle{plain}
\bibliography{example_paper}

\appendix

\section{Attention-Based Models}
\label{section:attention-models}
We review positional encoding in the context of Transformer models~\citep{DBLP:journals/corr/VaswaniSPUJGKP17}. The central building block of these models is multi-head attention and each attention head is calculated as follows:

\begin{equation}
\label{eq:attention}
    \mbox{Attention}(Q,K,V)=\mbox{softmax}(\frac{QK^{T}}{\sqrt{d_{k}}})V
\end{equation}

where queries $Q\in{\mathcal{R}^{N\times d_{k}}}$, keys $K\in{\mathcal{R}^{N\times d_{k}}}$, and values $V\in{\mathcal{R}^{N\times D_{v}}}$. $N$ is the number of items to consider, e.g., the number of tokens in a sequence or the number of pixel patches in an image. $d_{k}$ is the dimension of a key and query, and $D_{v}$ is the dimension of a value vector. Queries, keys and values are acquired via a linear projection of the input at each attention layer. For self-attention, they share the same input: 
\begin{equation}
\label{eq:qkv}
Q=E_{X}M_{Q}; 
K=E_{X}M_{K};
V=E_{X}M_{V}
\end{equation}
where $M_{Q}\in \mathcal{R}^{|E_{X}|\times d_{k}}$, $M_{K}\in \mathcal{R}^{|E_{X}|\times d_{k}}$ and $M_{V}\in \mathcal{R}^{|E_{X}|\times d_{v}}$ are the linear projection. $E_{X}\in \mathcal{R}^{N\times |E_{X}|}$ is the embedding of input $X$, which is jointly represented by its content embedding, $C_{X}$, and its positional encoding, $P_{X}$. 

\begin{equation}
\label{eq:embedding}
    E_{X}=C_{X} \oplus P_{X}
\end{equation}

where $\oplus$ can be either concatenation or element-wise addition. %$\mbox{Embed}(x)$ is then fed to the downstream computation that can be the attentional mechanisms at the input layer \citep{DBLP:journals/corr/VaswaniSPUJGKP17} or every hidden layer \citep{carion2020endtoend}. 
Previous work has investigated different combinations and decomposition of positional encoding and content embedding~\cite{he2021deberta}.
While concatenation and addition provide comparable results, the lack of positional encoding, $P_{X}$, will cause a significant drop in accuracy~\citep{DBLP:journals/corr/VaswaniSPUJGKP17,carion2020endtoend,li-etal-2020-widget}. In this paper, we investigate methods for realizing $P_{X}$. Note that for all the models except DETR, $P_X$ joins the content embedding as the input to the first layer. For DETR, $P_X$ is added to the input of every Transformer encoder layer, i.e., the activation of the previous Transformer layer.

%\section{Additional Results}
%In this section, we report additional ablation results.

%\subsection{Imagenet64 with Reformer}

%We further investigate the effect of using a non-linear transformation of MLP on several positional encoding methods (Figure~\ref{fig:reformer_ablation}). As we can see, using MLP generally improves the accuracy of each method. We also compare two top methods: Embed-2D + MLP and Learnable-Fourier + MLP. We found Learnable-Fourier + MLP shows faster convergence at the beginning, due to the inductive bias introduced by Fourier features. The two methods reach similar accuracy after 30k steps. However, it is important to mention that Embed-2D + MLP uses 50k more parameters than Learnable-Fourier + MLP.

%\begin{figure}
%\centering
%\includegraphics[width=0.94\textwidth]{figures/reformer_ablation.pdf}
%\caption{The impact of using MLP as a non-linear projection for each positional encoding method in the Imagenet64 Reformer experiments. The Y axis is the Bits per dim (bpd) and the X axis is training steps on the held-out data. The plot shows the mean of 3 repeats of experiments for each method.}
%\label{fig:reformer_ablation}
%\end{figure}

\section{Learned Positional Encoding Analysis}
\label{section:learned_features}

Our positional encoding is seeded with Fourier features whose dot product approximates $L_2$ distances---that brings the inductive bias to the model, which then evolves as learning progresses. %Both the weights $W_{r}$ and the target variance of these weights $\bar\sigma^{2}$ are trainable (see Equation 9 and Algorithm 1), which can change to approach an ideal Gaussian distribution that matches the data and the task. 
In this section, we analyze the positional encodings learned from the image generation, object detection and widget Captioning tasks. Note that the following analysis is focused on the output of Equation~\ref{eq:random_kernel} instead of that of Equation~\ref{eq:mlp}. The Fourier features directly represent the position while the MLP is trained to modulate the positional encoding to merge with the content embedding. It is less informative to analyze the MLP output because it neither directly represents the position nor directly participates in dot product attention (Equation~\ref{eq:attention}). In all the image benchmarks, the MLP output will be added to the content embedding and the addition is further processed by the transformation with $M_k$ in Equation~\ref{eq:qkv}. In the widget captioning benchmark, the MLP output will be concatenated with the content embedding and then projected by a dense layer to a hidden dimension required by the Transformer, which is further transformed by $M_k$ before dot product attention. 

% \begin{figure*}
% \centering
% \includegraphics[width=7in]{figures/reformer_heatmap.png}
% \caption{Reformer heatmap of similarity between top-left/center/bottom-right position and other positions.}
% \label{fig:learned_reformer}
% \end{figure*}

%\begin{figure*}
%\centering
%\includegraphics[width=7in]{figures/caption_embed4d_heatmap.png}
%\caption{Widget position heatmap of similarity between top-left/center/bottom-right position and other positions using embedding of the top and left coordinates of bounding box, trained as in \citet{li-etal-2020-widget}.}
%\label{fig:widget-heatmap}
%\end{figure*}

\subsection{PE Analysis for Image Generation Tasks}
\begin{figure}
\centering
\includegraphics[width=5.5in]{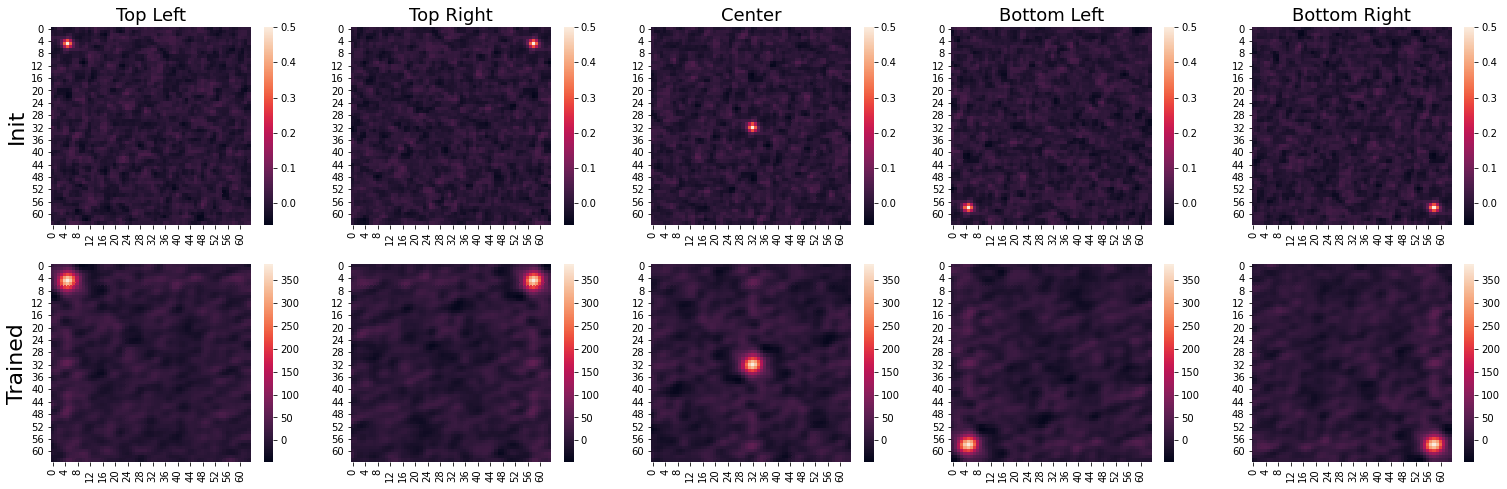}
\caption{The positional similarity, $r_{x}\cdot r_{y}$, of different positions on an image, to the rest of the positions on an image, as learned by Learnable-Fourier+MLP in Reformer. The Fourier features are initialized with weights drawn from a normal distribution: $\gamma=1.0$. The Top-Left, Top-Right, Center, Button-Left, and Bottom-Right positions are at (4, 4), (4, 57), (31, 31), (57, 4), (57, 57) on the image pixel grid.}
\label{fig:learned_reformer_1_mlp}
\end{figure}

\begin{figure}
\centering
\includegraphics[width=5.5in]{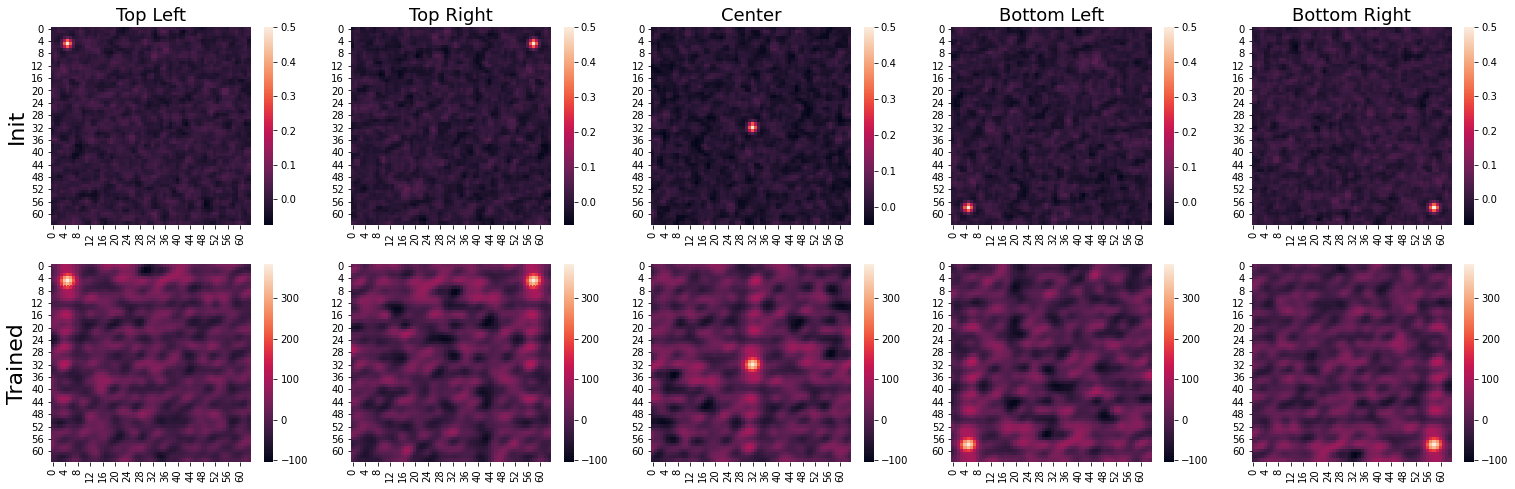}
\caption{The positional similarity learned by Learnable-Fourier \textit{without using the MLP modulator} in Reformer. The Fourier features are initialized with weights drawn from a normal distribution: $\gamma=1.0$.}
\label{fig:learned_reformer_1_na}
\end{figure}
Figure~\ref{fig:learned_reformer_1_mlp} visualizes the similarity of a given position on a $64\times 64$ image to the rest of the positions on the image, at the initial stage and the end of the training. The similarity is computed based on the dot product of the positional encoding of each position. The first row, Init, shows the similarity heatmap resulted from the initially seeded Fourier features based on $\gamma=1.0$. The second row, Trained, shows the similarity from the positional encoding learned after 100K steps when the model converges. As we can see, the positional relationship becomes less concentrated than the initialization, i.e., the "ball" becomes larger. %Based on this observation, we experimented with $\gamma=1$, which results in an even faster convergence, although it acquires a similar accuracy with $\gamma=64$ after 10k training steps. As discussed earlier, because $\bar\sigma^{2}$, which is initialized with $\gamma^{-2}$, is trainable, it gives the model flexibility to reach an ideal concentration about positional relationships by learning from the data. 
To further understand the impact of having the MLP modulator on the positional encoding, we compare the learned positional encoding with and without the MLP modulator. When there is no MLP modulator (Figure~\ref{fig:learned_reformer_1_na}), the learned positional encoding is less clean than the one with MLP. We suspect it is because without MLP, the positional encoding needs to directly participate in the addition with the content embedding (Equation~\ref{eq:embedding}). As a result, the encoding is not only learning to represent positions but also pressured to work with content embedding. As we show in our experiments, the lack of the MLP modulator results in a decrease in accuracy in this task.

\subsection{PE Analysis for Object Detection Tasks}

\begin{figure}
\centering
\includegraphics[width=5.5in]{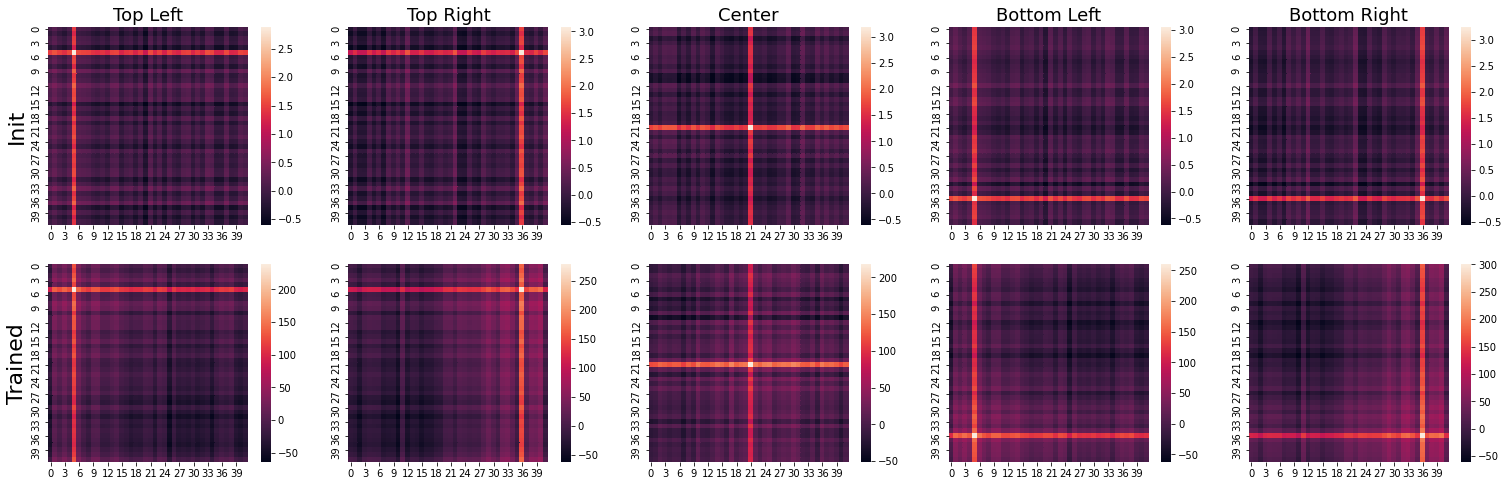}
\caption{Positional similarity visualization of Embed-2D positional encoding in DETR for object detection. %The Top-Left, Top-Right, Center, Button-Left, and Bottom-Right positions are at (5, 5), (5, 38), (21, 21), (38, 5), (38, 38) in the $42\times 42$ grid.
}
\label{fig:apd_detr_learned_emb}
\end{figure}

\begin{figure}
\centering
\includegraphics[width=5.5in]{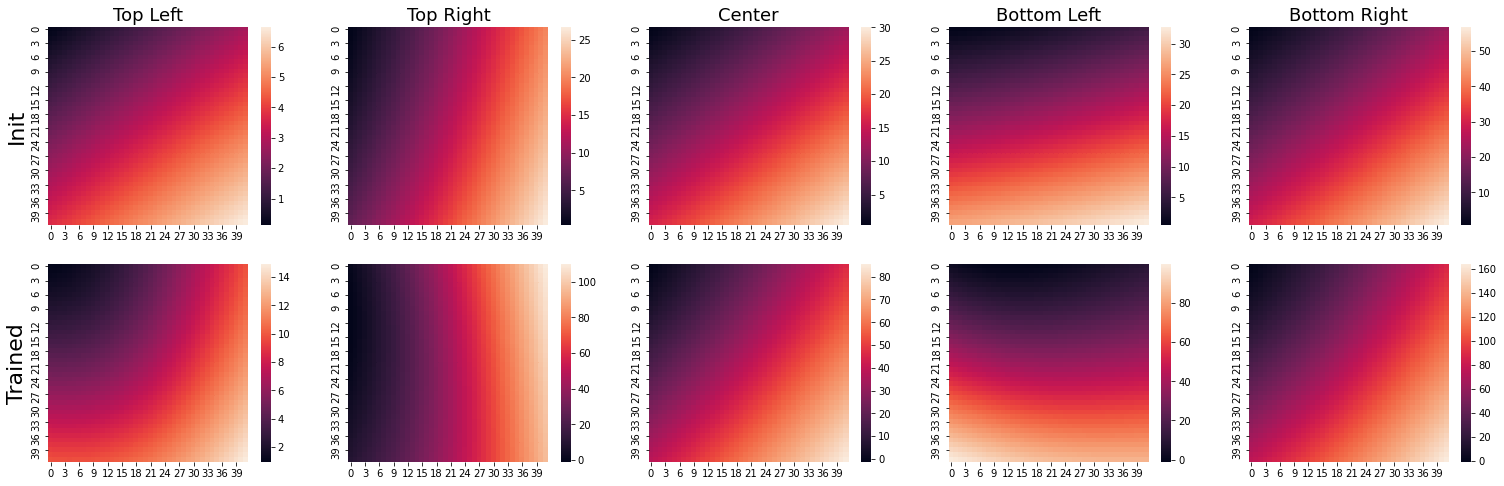}
\caption{Positional similarity visualization of MLP positional encoding in DETR for object detection.}
\label{fig:apd_detr_mlp}
\end{figure}

\begin{figure}
\centering
\includegraphics[width=5.5in]{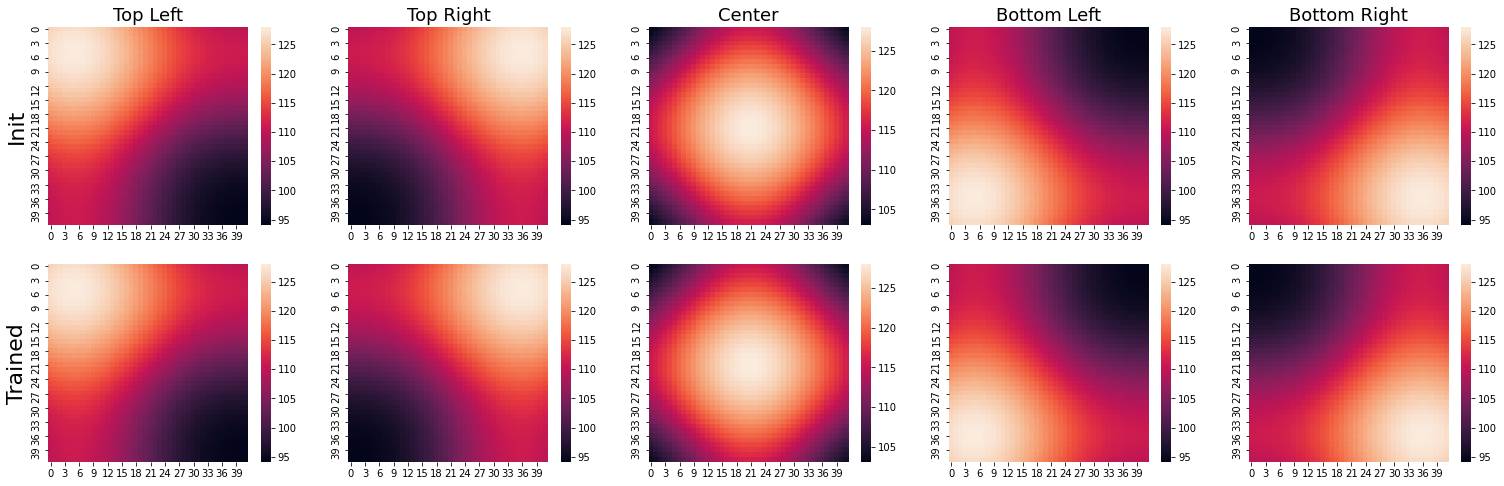}
\caption{Positional similarity visualization of Sine-2D positional encoding in DETR for object detection. The heatmap of the initial similarity and the "trained" similarity are the same because this method is parameter free.}
\label{fig:apd_detr_sine}
\end{figure}

\begin{figure}
\centering
\includegraphics[width=5.5in]{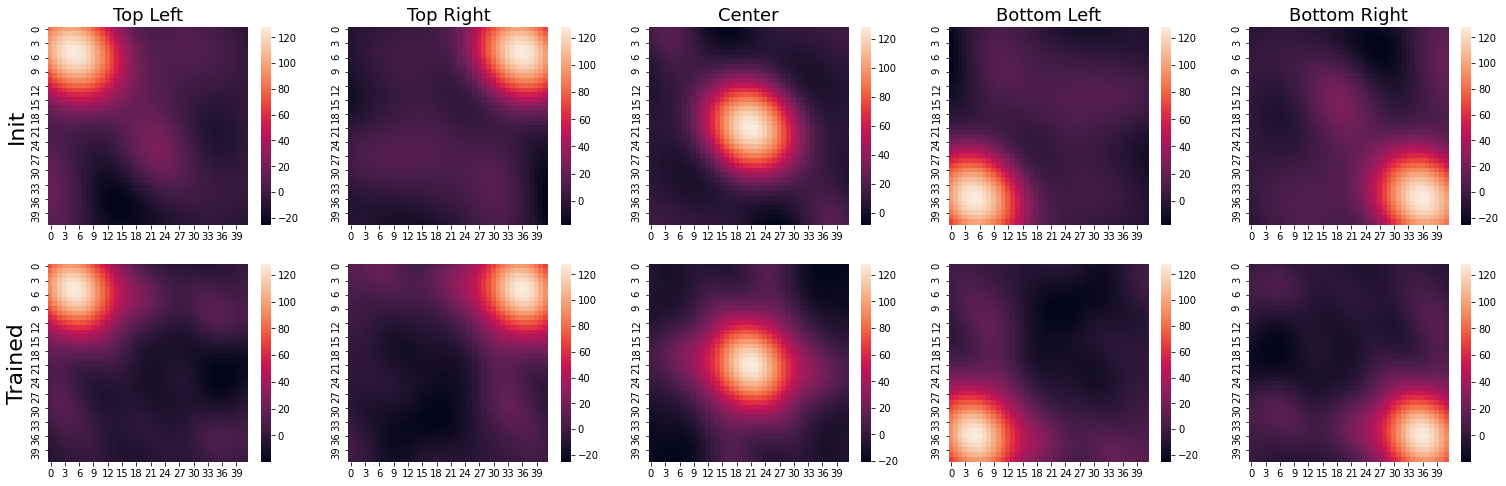}
\caption{Positional similarity visualization of Learnable-Fourier+MLP in DETR for object detection. The Fourier features are initialized with weights drawn from a normal distribution: $\gamma=1.0$.}
\label{fig:apd_detr_fourier}
\end{figure}

We visualize the initial and the learned positional relationships of each method in DETR for the object detection task (see Figures~\ref{fig:apd_detr_learned_emb}-\ref{fig:apd_detr_fourier}). Similar to the previous analysis, we analyze several representative positions on the $42\times 42$ grid in DETR, including the Top-Left (5, 5), Top-Right (5, 38), Center (21, 21), Button-Left (38, 5), and Bottom-Right (38, 38) positions, and the heatmaps show the positional similarity of these positions to the rest positions on the grid. 

By comparing the similarity heatmap of its initial and trained embedding weights (Figure~\ref{fig:apd_detr_learned_emb}), we found Embed-2D slowly learns spatial relationships between positions, as closer positions becomes more similar (brighter) in the heatmaps. Because the method concatenates independently embedded dimensions, it favors orthogonal directions like Sine-2D, as shown in Figure~\ref{fig:heatmap_concat}. 

The learned positional similarity of MLP (Figure~\ref{fig:apd_detr_mlp}) is skewed towards the bottom and the right directions based on the five analyzed positions. The heatmap intensity is based on the dot product of PEs, which is not normalized by their magnitudes like cosine similarities. As a result, the heatmap intensity towards the left and top edges is generally smaller (darker). Note that MLP does not have the shift-invariant property and the pattern of these five positions do not necessarily generalize across the entire grid space. 

For Sine-2D, its similarity heatmap obeys the "cross" pattern that we see in Figure~\ref{fig:heatmap_concat}. In DETR, position normalization allows positional encoding to concentrate on the center area of the cross (Figure~\ref{fig:apd_detr_sine}). As a result, the orthogonal bias is much reduced. Finally, we see Learnable-Fourier+MLP was able to mostly maintain ball-shaped similarity pattern throughout the training (Figure~\ref{fig:apd_detr_fourier}).

\subsection{PE Analysis for Widget Captioning Tasks}
\begin{figure}
\centering
\includegraphics[width=5.5in]{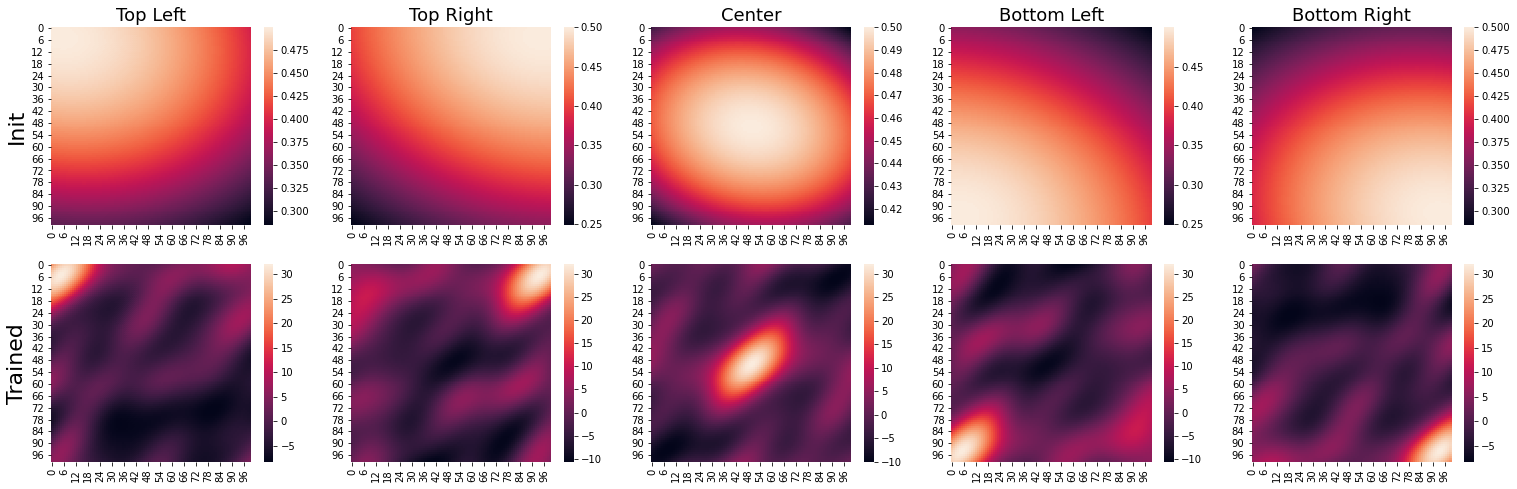}
\caption{The positional similarity of a UI screen, learned by Learnable-Fourier+MLP-2/2 for widget captioning. Note that in this task, each position is defined as a 4-coordinate bounding box. The heatmap only visualizes the point-wise similarity. The Fourier features are initialized with weights drawn from a normal distribution: $\gamma=100.0$.}
\label{fig:widget-heatmap}
\end{figure}

\begin{figure}
\centering
\includegraphics[width=5.5in]{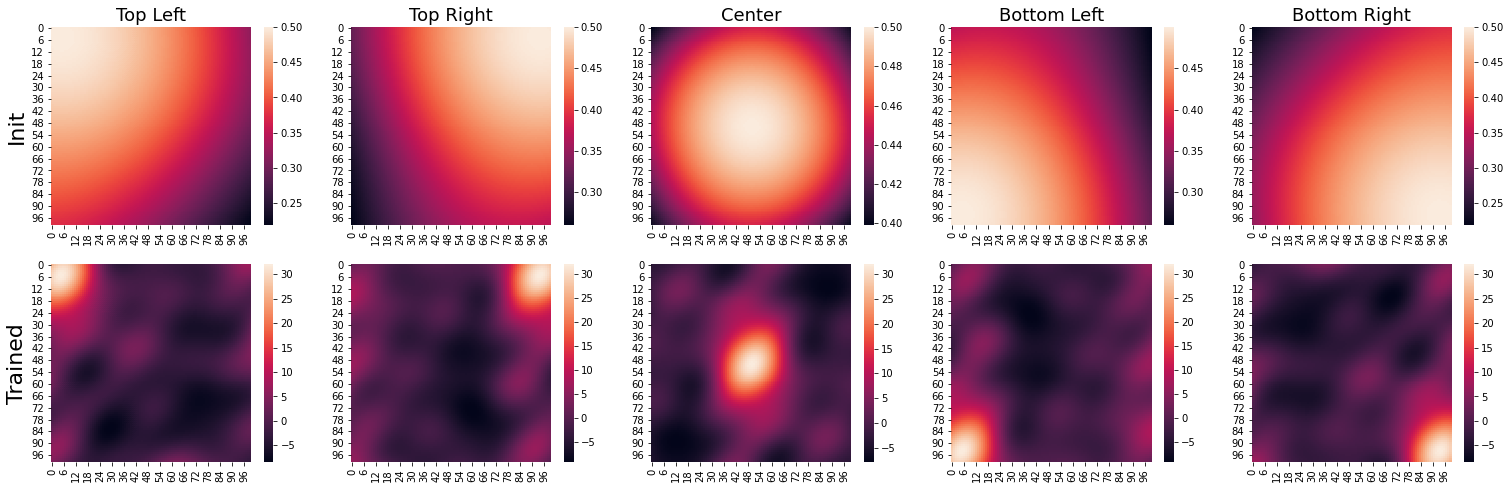}
\caption{The positional similarity of a UI screen, learned by Learnable-Fourier+MLP-2/2 with the KL loss (Equation~\ref{eq:kl} and \ref{eq:regularizer}) for widget captioning. The Fourier features are initialized with weights drawn from a normal distribution: $\gamma=100.0$.}
\label{fig:widget-heatmap_kl}
\end{figure}
Positional relationships are more complex in the widget captioning task, because each position is defined as a four-coordinate bounding box. We consider point-wise similarity a building block for bounding box similarity as discussed in the paper (Section~\ref{sec:widget_captioning}). Figure \ref{fig:widget-heatmap} shows the point-wise positional similarity learned by Learned-Fourier+MLP 2/2, which groups four coordinates into two groups to represent the top-left corner and the right-bottom corner positions of a bounding box. In this task, we see a more spread positional relationship than that of the image generation task, because we seed the Fourier features with $\gamma=100$ in this task. We observed that the positional relation becomes more concentrated over the course of the training than that of the initial encodings. We also see the positional relation distribution becomes more skewed (towards the anti-diagonal direction). To understand whether maintain the symmetry of the distribution would help on accuracy, we conduct additional experiments by applying a regularizer to the Fourier weights $W_r$ as the follow.

\begin{equation}
\label{eq:kl}
    \Lagr_{KL}=-\frac{1}{2}(1-\log\bar{\sigma}^{2}+\log\sigma^{2}-\frac{\sigma^{2}+\mu^{2}}{\bar{\sigma}^{2}})
\end{equation}

where $\mu$ and $\sigma^{2}$ are the mean and variance of $W_{r}$. $\bar{\sigma}^{2}$ is the target variance that is also learnable, which is initialized as 
$\gamma^{-2}$. The KL loss ensures $W_{r}$ to obey a Gaussian distribution centered at $0$ thus maintains the symmetry of positional relationships along all the directions. When training the model, the regularizer loss $\Lagr_{KL}$ is added to the overall loss for optimization.
\begin{equation}
\label{eq:regularizer}
    \Lagr_{total}=\Lagr_{model}+\alpha \Lagr_{KL}
\end{equation}
\vspace{-1pt}

\begin{figure}[ht]
\centering
\subfigure[Initial distribution of $W_r$.]{%
\includegraphics[height=1.1in]{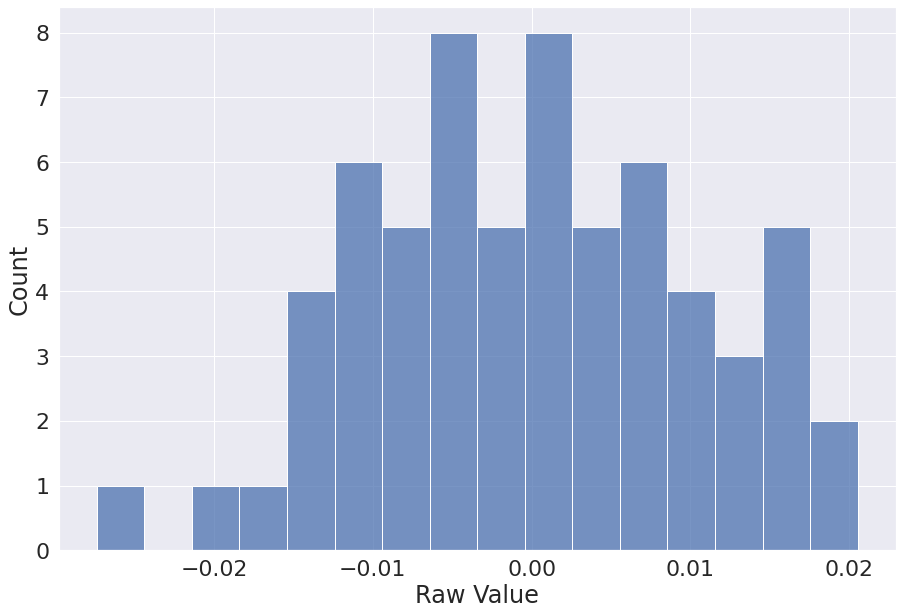}
\label{fig:kernel_2_init_hist}}%
\qquad
\subfigure[$W_r$ learned w/o the KL loss.]{%
\includegraphics[height=1.1in]{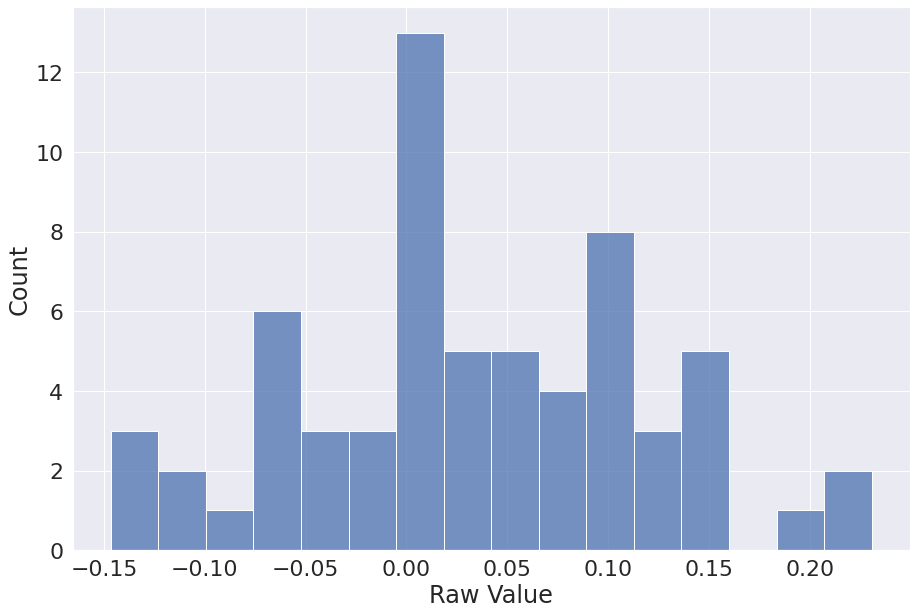}\label{fig:kernel_2_no_klloss_hist}}%
\qquad
\subfigure[$W_r$ learned with the KL loss.]{%
\includegraphics[height=1.1in]{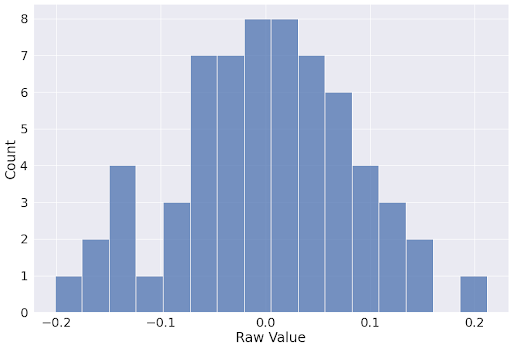}\label{fig:kernel_2_klloss_hist}}%
\caption{The distribution of $W_r$.}
\label{fig:w_histogram}
\end{figure}

In this experiment, we use $\alpha=1$. The resulted positional encoding is shown in Figure~\ref{fig:widget-heatmap_kl}. As we can see, the symmetry of the positional relation distribution is better maintained with the KL loss, and the distributions of initial and learned $W_r$ for without and with the KL loss are shown in Figure~\ref{fig:w_histogram}. We see a clear improvement of accuracy with the use of this KL loss for Learned-Fourier+MLP 2/2. However, using the KL loss does not seem to impact image-based tasks much, e.g., image generation and object detection tasks. We suspect that as shown in Figure~\ref{fig:learned_reformer_1_mlp}, the symmetry of positional relation distribution is naturally maintained even without using the KL loss. Thus KL loss is less useful in such cases.

\section{Additional Ablation Studies}
\label{sec:ablation}

%We compare our positional encoding based on learnable Fourier versus constant Fourier features, both with and without an additional MLP layer. We found constant Fourier features alone (\textit{Fourier}) do not perform as well as those with learnable parameters (Figure~\ref{fig:reformer_fourier}). When it is enhanced with an additional MLP layer (\textit{Fourier + MLP}), it performs nicely. Overall, Learnable-Fourier + MLP still performs the best. These experiments show that learnability is crucial for our methods to work.

It is possible to extend traditional sinusoidal positional encoding (Equation~\ref{eq:sin-cos}) for the multi-dimensional positions by using multi-dimensional frequencies, instead of using the concatenation of independently encoded spatial dimensions. For 2D positions on an image, we can linearly combine the vertical and horizontal positions using constant frequencies that are manually determined. In this ablation, we adapt the original Transformer sinusoidal frequencies for each dimension. Specifically, for a 2D position $(x, y)$, the multi-dimensional sinusoidal PE, referred as \textit{Transformer MD-Sine}, is the follow, where $D$ is the dimension of the PE and $0\leq d\leq \frac{D}{2}$. 

\begin{equation*}
        PE(p,2d) = \sin{(\frac{x}{10000^{2d/D}} +\frac{y}{5000^{2d/D}})}; 
        PE(p,2d+1) = \cos{(\frac{x}{10000^{2d/D}} +\frac{y}{5000^{2d/D}})}
\end{equation*} 

\begin{figure}[t]
\centering
\includegraphics[height=2in]{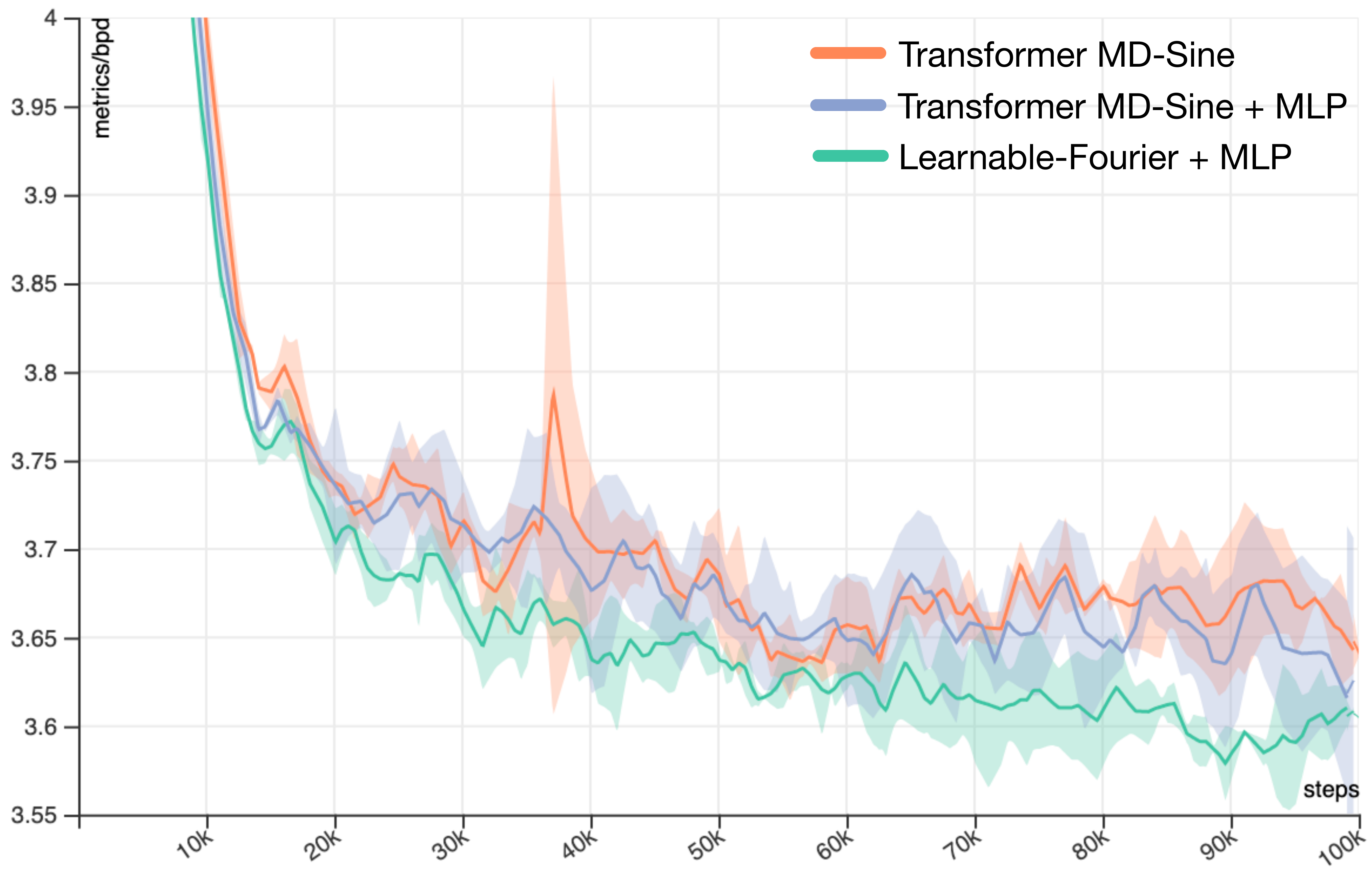}
\caption{Bits per dim (bpd) w.r.t. training steps on the image generation task with Reformer. The ablation compares learnable Fourier features with multi-dimensional sinusoidal PE based on Transformer frequencies. The plot shows the mean and 95\% confidence interval based on 3 repeats of experiments for each method.}
\label{fig:appendix_mdsine}
\end{figure}

As shown in Figure~\ref{fig:appendix_mdsine}, Transformer MD-Sine performs poorly in the Reformer Imagenet64 task. Adding MLP to Transformer MD-Since improves its performance, but it still does not perform as good as Learnable Fourier. Although it is possible to find better constant frequencies for linearly combining these dimensions, it can be effort consuming to manually tune these frequencies to perform optimally. In contrast, our approach with learnable Fourier features lets the model learn these frequencies that are appropriate for the task.

\begin{table}[ht!]
\caption{The performance of Sine-4D when it is enhanced by an MLP for the widget captioning task.}
\label{tab:sine4d-mlp}
\centering
\begin{tabular}{lcccccc}
    \toprule
    {Method}
    & {BLEU-1}
    & {BLEU-2}
    & {ROUGE}
    & {CIDEr}
    & {METOER}
    & {SPICE} \\
    \midrule
    Sine-4D & 44.9 & 31.9 & 43.9 & 94.9 & 31.0 & 16.7 \\
    Sine-4D+MLP-1/4 & 45.3 & \bf 32.4 & 45.0 & 97.6 & 31.9 & 16.9 \\
    Sine-4D+MLP-2/2 & \bf 45.4 & 32.1 & \bf 45.2 & \bf 98.1 & \bf 32.0 & 17.3 \\
    Sine-4D+MLP-4/1 & 45.3 & 32.3 & 44.8 & 97.5 & 31.9 & \bf 17.7 \\
   \bottomrule
\end{tabular}
\end{table}

We found MLP is often beneficial when it is added to an existing positional encoding such as sinusoidal or embedding based methods. For example, the overall accuracy of Sine-4D is improved when an MLP is used with it for the widget captioning task (Table~\ref{tab:sine4d-mlp}). For certain tasks, a dense transform or even simpler scaling over Fourier features (Equation~\ref{eq:random_kernel}) can lead to good results, e.g., the object detection task. Yet, using an MLP seems to consistently offer good results across tasks.

\begin{figure}
\centering
\includegraphics[width=5in]{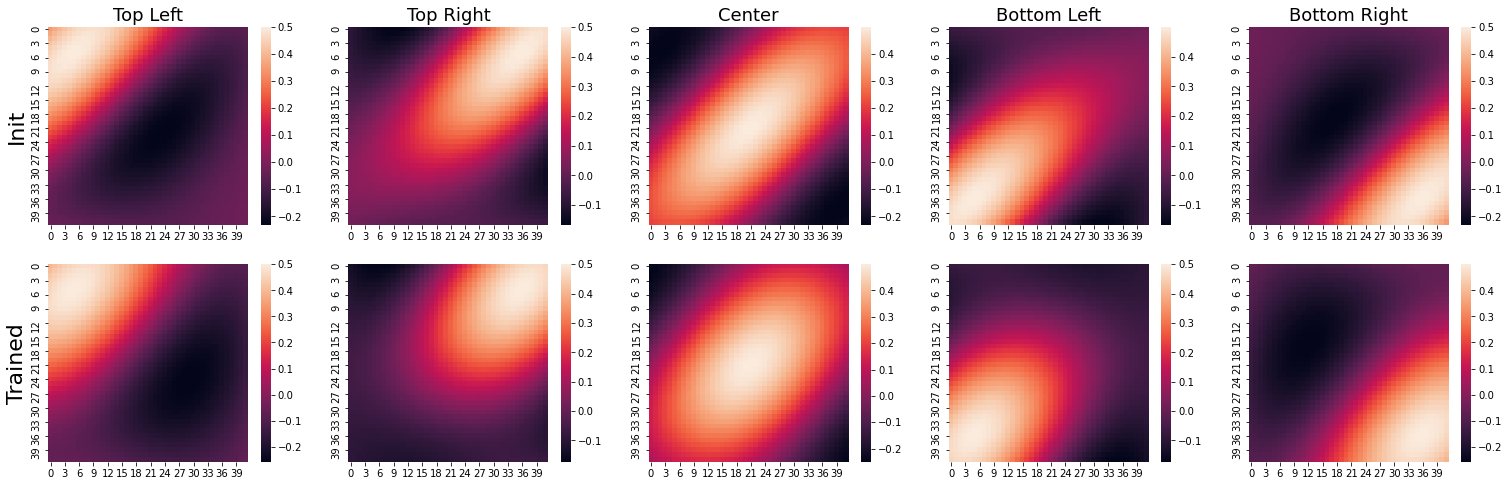}
\caption{The initial and learned positional relationships of Fourier features when $W_r$ is initialized by drawing from a uniform distribution $[0,1]$. The Top-Left, Top-Right, Center, Button-Left, and Bottom-Right positions are at (5, 5), (5, 38), (21, 21), (38, 5), (38, 38) in the $42\times 42$ grid. The Fourier features are initialized with weights drawn from a normal distribution: $\gamma=4.0$.}
\label{fig:apd-detr-uniform}
\end{figure}

Finally, we compare the performance our positional encoding when $W_r$ is initialized from a different distribution. In this ablation, we initialize $W_r$ by drawing from a uniform distribution in the range of $[0, 1]$ in comparison with drawing from a normal distribution (Equation~\ref{eq:guassian}). In the object detection task (Table~\ref{tab:dist}), initializing $W_r$ from the uniform distribution performs worse than from a normal distribution. When the learnable Fourier features are enhanced by the MLP layers, the performance of using both initialization distributions are improved and reach a similar level of performance, although drawing from the normal distribution still has a slight advantage. By examining the learned Fourier features from uniform initialization, we found the positional relationships, as visualized by the heatmaps, has become more "round" or towards a ball shape after learning than those at the initialization (Figure~\ref{fig:apd-detr-uniform}), which indicates that the model is more inclined to $L_2$ distances between positions.
%and its $W_r$ still obeys a normal distribution (Figure~\ref{fig:apd_detr_fourier_weight})

\begin{table}
  \caption{Performance for initializing $W_r$ with different distributions, and with and without MLP.}
  \label{tab:dist}
  \centering
  \begin{tabular}{lllllll}
    \toprule
    Configuration     & $AP$     & $AP_{50}$ & $AP_{75}$     & $AP_{small}$ & $AP_{medium}$ & $AP_{large}$\\
    \midrule
    Uniform [0, 1]     & 38.3 & 59.4 & 40.0 & 17.7 & 41.9 & 56.9      \\
    Uniform [0, 1] \& MLP     & 40.0 & 60.5 & 42.0 & 18.4 & \bf 43.5       & 58.9  \\
    $\mathcal{N}(0,\,4^{-2})$     & 39.1       & 60.0 & 40.9 & 18.1 & 42.5 & 58.0  \\
     $\mathcal{N}(0,\,4^{-2})$ \& MLP & \bf 40.2  & \bf 60.7 & \bf 42.4 & \bf 20.0 & 43.3 & \bf 59.0     \\
    \bottomrule
  \end{tabular}
\end{table}

%\begin{figure}
%\centering
%\includegraphics[width=2.6in]{figures/sinemd_weight.png}
%\caption{The histogram of $W_r$ of Learnable-Fourier+MLP in DETR when the model converges for the object detection task.}
%\label{fig:apd_detr_fourier_weight}
%\end{figure}

\section{Hyperparameters \& Parameter Sizes}
\label{section:param_sizes}
%In this section, we add more details about the hyperparameters and parameter sizes of each model in our experiment. %We use $\alpha=1.0$ for the KL-divergence regularizer loss for all the experiments (see Equation 10).

\begin{table}[h]
  \caption{The model parameter sizes of Reformer~\citep{kitaev2020reformer} with different positional encoding methods.}
  \label{tab:reformer_size}
\centering
\begin{tabular}{lc}
    \toprule
    {Method}
    & {Reformer Model Parameter Size} \\
    \midrule
    Embed-1D & 73.2M \\
    Embed-2D & 60.7M \\
    %Embed-2D + MLP & 60.72M \\
    Sine-1D & 60.6M \\
    Sine-2D & 60.6M \\
    Transformer MD-Sine & 60.6M \\
    Transformer MD-Sine + MLP & 60.7M \\
    %Sine-2D + MLP & 60.67M \\
    MLP & 60.6M \\
    %MLP + MLP & 60.7M \\
    %Fourier & 60.6M \\
    %Fourier + MLP & 60.7M \\
    Learnable-Fourier & 60.6M \\
    Learnable-Fourier + MLP & 60.7M \\
    \bottomrule
\end{tabular}
\end{table}

\begin{table}[h]
  \caption{The model parameter sizes of DETR~\cite{carion2020endtoend} with different positional encoding methods.}
  \label{tab:detr_size}
\centering
\begin{tabular}{lc}
    \toprule
    {Method}
    & {DETR Model Parameter Size} \\
    \midrule
    Embed-2D & 41.6M \\
    Sine-2D & 41.6M \\
    MLP & 41.6M \\
    Learnable-Fourier + MLP & 41.6M \\
    \bottomrule
\end{tabular}
\end{table}

\begin{table}[h!]
  \caption{The model parameter sizes of the widget captioning model~\citep{li-etal-2020-widget} with different positional encoding methods.}
  \label{tab:caption_size}
\centering
\begin{tabular}{lc}
    \toprule
    {Method}
    & {Widget Captioning Model Parameter Size} \\
    \midrule
    SOTA~\cite{li-etal-2020-widget} & 5.11M  \\
    Embed-4D & 5.11M \\
    MLP & 5.08M \\
    Sine-4D & 5.07M \\
    Sine-4D+MLP-1/4 & 5.07M \\
    Sine-4D+MLP-2/2 & 5.07M \\
    Sine-4D+MLP-4/1 & 5.08M \\
    Learnable-Fourier-2/2 & 5.07M \\
    Fixed-Fourier+MLP-1/4 & 5.10M \\
    Fixed-Fourier+MLP-2/2 & 5.08M \\
    Fixed-Fourier+MLP-4/1 & 5.07M \\
    Learnable-Fourier+MLP-1/4 &  5.11M \\
    Learnable-Fourier+MLP-2/2 & 5.07M \\
    Learnable-Fourier+MLP-4/1 & 5.07M \\
   \bottomrule
\end{tabular}
\end{table}

For Reformer experiments, each model is based on the Reformer model for the Imagenet64 task~\citep{kitaev2020reformer}. The number of parameters for each Reformer model is summarized in Table~\ref{tab:reformer_size}. We here focus on the positional encoding part of the model that is where each variant differs. Our positional encoding, Learnable-Fourier+MLP, uses roughly the same number of trainable parameters as Embed-2D, the benchmark method used in the original Reformer. All the Fourier-based methods used $|F|=768$, $|H|=32$, $D=768$ and $\gamma=1.0$. For the MLP, we used LayerNorm before each dense projection, $W_1$ and $W_2$ (see Algorithm~\ref{alg:psedo1}). We set $G=1$ because vertical and horizontal positions need to be mapped jointly to model the inductive bias of $L_2$ distances on an image. Embed-1D uses significantly more parameters because it needs to assign an embedding vector for each position in a flattened image. Sine-1D and Sine-2D are parameter-free encoding, thus use the least parameters. %We follow the experimental procedure as detailed in the Reformer paper. All our experiments used a 6-layer, 8-head-attention Reformer, with $d_{model} = 1024$, $d_{ff} = 4096$, and $n_{heads} = 8$. These models are implemented based on the public Reformer codebase in Trax\footnote{\url{https://github.com/google/trax/tree/master/trax/models/reformer}}. The training for each Reformer model is parallelized across 32 TPU v2 cores, and each batch contains 8 sequences (images) on each core. We trained each model variant for 100k steps, which took about 24 hours to complete.

The parameter sizes for each DETR model~\citep{carion2020endtoend} are shown in Table~\ref{tab:detr_size}. All the variants of DETR roughly uses the same number of trainable parameters. We used $\gamma=1.0$ for Learnable-Fourier + MLP in Section~\ref{section:detr}. The MLP uses a dense layer $2\times 256$ with GeLU as activation. %We did not use any dropout in these methods. %For Learnable-Fourier+MLP in the DETR experiments, we did not use MLP after the learnable random features $r_{x}$: $|F|=256$, $\gamma=0.5$ and $G=1$. %All our experiments with Vision Transformer are based on ViT-Base. See~\citep{visiontransformer} for model architecture details. In particular, $|F|=|H|=D=768$  and $G=1$, and $\gamma$ is the grid size (either 28 or 14). The parameter sizes for each ViT model is shown in Table~\ref{tab:vit_size}.
%We use the default 6-layer Encoder-Decoder setup in DETR, using all the same hyperparameters, which uses COCO 2017 detection dataset with 118k images for training and 5k for validation. All the variants are based on the DETR codebase\footnote{\url{https://github.com/facebookresearch/detr/blob/master/models}}, which are ported into JAX\footnote{\url{https://github.com/google/jax}}, a library for machine learning. The training for each DETR model is parallelized across 64 TPU v3 cores with a batch size of 64 images. We follow the experimental procedure of the DETR paper. We trained each variant for 90k steps, which took roughly 16 hours to complete.

For UI widget Captioning experiments, the number of parameters of each model variant is shown in Table~\ref{tab:caption_size}. The model architecture that is shared by each model variant is summarized in the paper and detailed in the previous paper~\citep{li-etal-2020-widget}. For Fourier-based methods, we used $|F|=128, 64, 32$, $G=1, 2, 4$ for position grouping variants: 1/4, 2/2 and 4/1, respectively. We used $\gamma=100$ for initializing $W_{r}$ for all the Fourier-based methods. We used a dropout of 20\% after the non-linear activation in the MLP modulator. %We use the same model architecture and hyperparameters of the strongest model, \textit{Pixel+Local+Context}, as the original paper \citep{li-etal-2020-widget}, and built our experiment based on the public codebase of widget captioning\footnote{\url{https://github.com/google-research/google-research/tree/master/widget_caption}}. Specifically, the screen encoder uses a 6-layer, 8-head Transformer with a hidden size of 128. We train all the models to 100k steps with Adam optimizer and a scheduled learning rate detailed the original paper. All the models converged within 12 hours using 4 NVIDIA V100 GPU.

For computational complexity, embedding-based approaches generally require less computation than others. For the Reformer experiments, Embed2D, trains at 1.86 steps/second, Sine2D in contrast trains at 1.81 steps/second, and our Learnable-Fourier+MLP is slower and trains at 1.22 steps/second. For the experiments with the object detection and widget captioning tasks, the impact of using different PE methods on runtime is negligible because the rest computation in the training is more dominant, e.g., Hungraian matching for computing the minimal loss in DETR for object detection.

\section{Unseen Position Distribution in the Widget Captioning Dataset}

\begin{figure}
\centering
\includegraphics[width=3in]{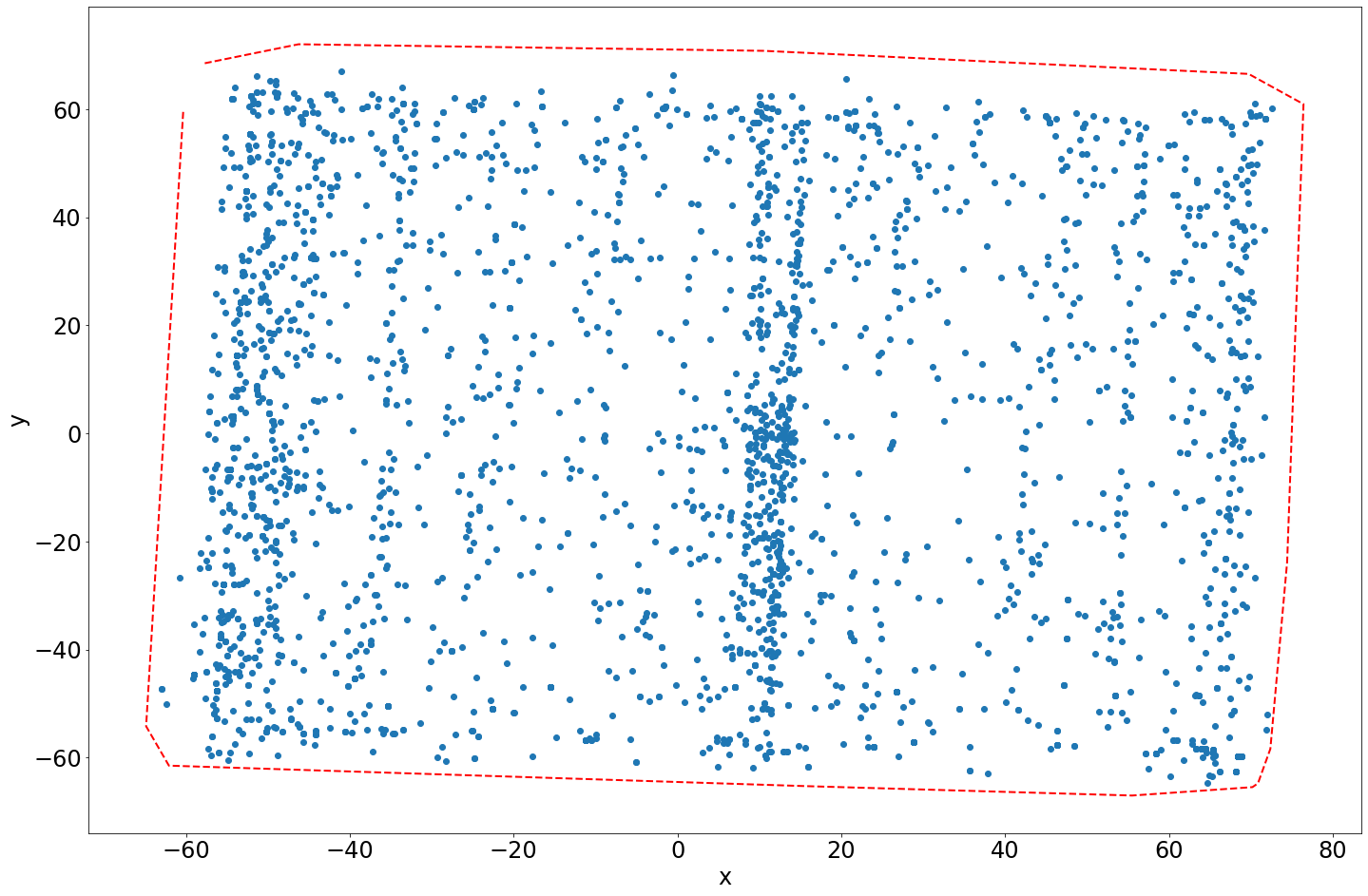}
\caption{The unseen positions in the test set within the convex hull of all the positions in the training set of the widget captioning dataset.}
\label{fig:widget-hull}
\end{figure}

One advantage of the proposed positional encoding method is to generalize to unseen positions. Our experiments with object detection include unseen positions that require "extrapolation". For the widget captioning task, we found 2685 of the 2692 unseen positions in the test dataset are inside the convex hull of all the training positions. To show the distribution of the unseen positions in the test set, we map all these unseen positions to 2D with PCA (see Figure \ref{fig:widget-hull}). We  plot the 2D convex hull of all the positions in the training set, which is also mapped via PCA, as the red dashed line. We can see all the unseen positions are within the convex hull of the training positions for this task.

\end{document}